\documentclass[10pt,table]{article} 
\usepackage[preprint]{tmlr}

\usepackage{amsmath,amsfonts,bm}

\def\eqref#1{equation~\ref{#1}}

\def\1{\bm{1}}

\def\vb{{\bm{b}}}

\def\vg{{\bm{g}}}

\def\vm{{\bm{m}}}

\def\mA{{\bm{A}}}

\def\mU{{\bm{U}}}
\def\mV{{\bm{V}}}

\DeclareMathAlphabet{\mathsfit}{\encodingdefault}{\sfdefault}{m}{sl}
\SetMathAlphabet{\mathsfit}{bold}{\encodingdefault}{\sfdefault}{bx}{n}

\newcommand{\R}{\mathbb{R}}

\DeclareMathOperator{\sign}{sign}

\usepackage{hyperref}
\usepackage{url}

\usepackage{graphicx}
\usepackage{caption}
\usepackage{algorithm}
\usepackage{algpseudocode}
\usepackage{xcolor}
\usepackage{placeins}
\usepackage{multicol}
\usepackage{multirow}
\usepackage{cleveref}
\usepackage{subcaption}
\usepackage{lineno}
\usepackage{xspace}

\makeatletter
\DeclareRobustCommand\onedot{\futurelet\@let@token\@onedot}
\def\@onedot{\ifx\@let@token.\else.\null\fi\xspace}

 \def\Eg{\emph{E.g}\onedot}
\newcommand{\lmo}{\operatorname{lmo}}
\newcommand{\RMS}{\mathrm{RMS}}

\title{You May Be Running the Wrong Inception Crop}

\author{\name Jason Chuan-Chih Chou \email chuanchih@gmail.com\\
      \addr Cohere Labs Community}

\def\month{MM}  
\def\year{YYYY} 
\def\openreview{\url{https://openreview.net/forum?id=XXXX}} 

\begin{document}

\maketitle

\begin{abstract}
A decade after its inception, Inception crop has become the standard crop-based data augmentation method for training deep vision models. Not only is its practice of uniformly sampling crop scale and aspect ratio widely adopted, but also its lower and upper bounds, with the scale lower bound being the sole exception that is sometimes tuned. It is therefore surprising that the standard implementation in the TensorFlow / JAX ecosystem samples crop scale with probability density function $f(A) \propto \frac{1}{\sqrt{A}}$ unlike the PyTorch counterpart, which follows the original description. Motivated by this discovery, we train 522 ViT-S/16 models on the ImageNet-1k dataset with various training budgets and crop scale distributions. We reach $78.78\pm0.09$ top-1 val. accuracy with 90 epochs of training budget and find that 1. Higher training budget requires stronger augmentation; 2. Lower tail of the distribution of the crop scale determines the augmentation strength of Inception crop; 3. Models trained with higher training budget exhibit sparser saliency, regardless of the crop scale distribution or weight decay. Based on 2. we revisit the performance of Beta crop, whose softer cutoff allows it to optimize model performance across training budgets with less compromise. We replicate 1. and 3. with Scion optimizer in addition to AdamW, suggesting that the results may be general.
\end{abstract}
\section{Introduction}
\label{sec:intro}
Originally proposed as a data augmentation method for training the Inception network \citep{7298594} on the ImageNet-1k dataset \citep{5206848}, Inception crop has become the de facto standard crop-based augmentation method for training deep vision models based on CNN (convolutional neural network) \citep{10.5555/3666122.3666890, wightman2021resnet,10.1007/978-3-031-73661-2_5} or ViT (vision transformer) \citep{beyer2022better,steiner2022how}, trained on smaller \citep{Jordan202494OC} or larger \citep{steiner2022how} datasets, for both supervised learning and self-supervised learning \citep{10.5555/3524938.3525087, chen2020big} models. Its practice of sampling crop scale\footnote{Crop area normalized s.t. area of the original image is 1} and aspect ratio\footnote{Width of the crop divided by its height} uniformly is usually taken for granted, and so are the range of the aspect ratio and the upper bound of the crop scale. The lower bound of the crop scale $a_\mathrm{min}$ is perhaps the only parameter considered tunable \citep{10.5555/3666122.3666890}, with smaller value resulting in stronger augmentation. It is therefore surprising when we track down the training loss discrepancy in reproducing the results from \citet{beyer2022better} to the TensorFlow implementation of Inception crop with an otherwise identical setup (\cref{sec:training_loss}). Given its popularity in training deep vision models in the TensorFlow / JAX ecosystem \citep{10.5555/3524938.3525087,chen2020big,zhai2022scaling,steiner2022how,beyer2022better,beyer2023flexivit,dehghani2024patch,alabdulmohsin2024getting,Dahl2023AlgoPerf}, we find it critical to assess its potential impact.

Prior work has experimented on how the strength of data augmentation may impact model performance with the crop scale lower bound of the Inception crop as the main focus \citep{10.5555/3666122.3666890} but experiments on the distribution of crop scale are more limited \citep{NEURIPS2021_a2fe8c05}. Furthermore, both are done in isolation from each other and neither explores how varying training budget (in terms of the number of training epochs) may affect the results. It is therefore imperative to assess such implementation differences in the full context of training budget and crop scale distribution. After extensive experiments of sensible combinations of these factors, we find that:

\begin{enumerate}
  \item Optimal augmentation strength depends on the training budget. Higher training budget requires stronger data augmentation.
  \item Lower tail of the distribution of the crop scale determines the augmentation strength of Inception crop.
  \item Models trained with higher training budgets exhibit sparser saliency which is unaffected by crop scale distribution or weight decay, regardless of how we measure it.
\end{enumerate}

Based on 2. we revisit the performance of Beta crop, whose softer cutoff allows it to optimize model performance across training budgets with less compromise therefore partially mitigates 1.
\section{Related Works}
\label{sec:related}

\subsection{ImageNet datasets and evaluation}

When introduced in 2012 for the ImageNet Large Scale
Visual Recognition Challenge (ILSVRC) \citep{russakovsky2015imagenet}, the ImageNet-1k dataset with its 1000 object classes and 1,461,406 images represented orders of magnitude jump in terms of the scale of labeled datasets. Partially due to the inaccessibility of the ImageNet-21k superset \citep{ridnik2021imagenetk}, the ImageNet-1k dataset remains the standard dataset for training and evaluating deep vision models. Nonetheless, ImageNet-1k is not without its flaws \citep{kisel2024flaws} such as faulty labels \citep{vasudevan2022when} and multi-class images \citep{9577610}. To mitigate the latter, \citet{beyer2020we} employ a hybrid approach using an ensemble of models and human annotators to create multi-label Reassessed Labels (``ReaL'') and use them to re-evaluate progress of the field.

\subsection{ViT training}

\citet{beyer2022better} significantly improve upon the ViT-S/16 ImageNet-1k baseline of the original ViT paper \citep{dosovitskiy2021an} and we use their result as the baseline in our study. Beyond the scale of ImageNet-1k, \citet{steiner2022how} conduct extensive experiments and find that ``AugReg'' (data augmentation and model regularization) tends to benefit ViT training on smaller datasets. The reported effect of AugReg on ImageNet-21k training for 30ep vs. 300ep further hints at the general results, but \citet{steiner2022how} do not report any experiments with the parameters of Inception crop.

\subsection{Data augmentation}
Data augmentation has a long history in the field of computer vision \citep{NIPS1991_65658fde} and has been considered a form of implicit regularization of the model \citep{hernandez-garcia2020data}. In addition to crop-based augmentation, transformation-based augmentation such as RandAugment \citep{NEURIPS2020_d85b63ef} and TrivialAugment \citep{9711158} and mixing-based augmentation such as Mixup \citep{zhang2018mixup}, CutMix \citep{9008296}, and AugMix \citep{hendrycks*2020augmix} have all been used alone or in combination to train deep vision models. Strong data augmentation, however, can lead to confusion between co-occurring or semantically similar classes due to degradation of the semantic meaning of the image \citep{10.5555/3666122.3666890}.

\subsection{Saliency map of deep vision models}
Saliency map refers to a heatmap that highlights the most relevant regions of the image that aims to make the model's inference result more explainable. For CNNs with ReLU activation function, the local connectivity pattern and ReLU's simplicity allow simple approaches such as guided backpropagation \citep{springenberg2015striving}, CAM \citep{zhou2016learning}, and Grad-CAM \citep{selvaraju2017grad} to generate intuitive results. Transformer models' all-pair attention mechanism and activation functions with both positive and negative regions turn out to be more challenging for these techniques, though methods such as LRP \citep{binder2016layer} and propagated relevance map \citep{chefer2021transformer} have shown some promise.

\section{Inception Crop Implementations}
\label{sec:inception}
\subsection{TensorFlow}
\begin{figure}[t]
  \centering
  \includegraphics[width=0.4\linewidth]{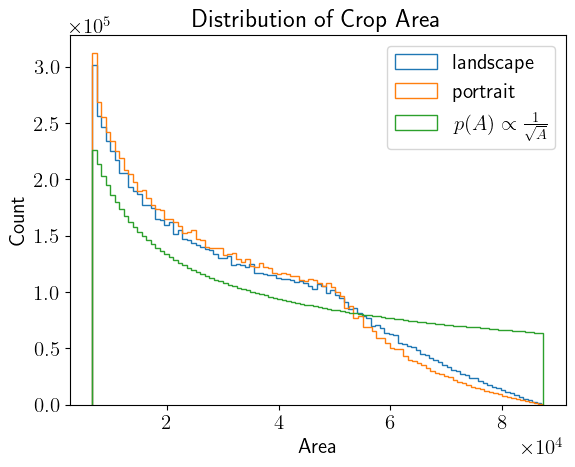}
  \caption{Crop area distribution of the standard implementation of Inception crop in TensorFlow with default parameters for $256 \times 512$ and $512 \times 256$ images, $N=10^7$. The probability density function of the distribution is approximately proportional to $\frac{1}{\sqrt{A}}$ but truncated at larger area when some aspect ratios become inadmissible. Note that landscape and portrait images have slightly different distributions.}
  \label{fig:tf_area_distribution_approx}
\end{figure}

\begin{figure}[t]
    \begin{subfigure}{\columnwidth}
    \centering
    \includegraphics[width=0.6\columnwidth]{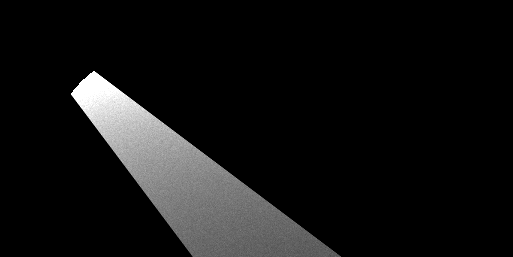}
    \caption*{Landscape image ($h_o=256, w_o=512$)}
    \end{subfigure}
    \begin{subfigure}{\columnwidth}
    \centering
    \includegraphics[width=0.6\columnwidth]{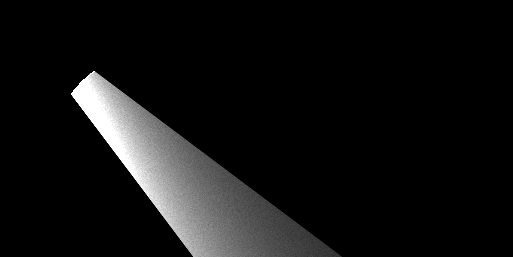}
    \caption*{Portrait image ($h_o=512, w_o=256$), transposed}
    \end{subfigure}%
    \caption{Probability heatmaps of the crop size $(h, w)$ given by the TensorFlow implementation for landscape and portrait images with default parameters $r_\mathrm{min} = \frac{3}{4}, r_\mathrm{max} = \frac{4}{3}, a_\mathrm{min} = 0.05, a_\mathrm{max} = 1, N=10^7$.}
    \label{fig:tf_heat_map}
\end{figure}

Standard implementation of Inception crop in TensorFlow that JAX vision libraries often depend upon \citep{big_vision,dehghani2021scenic} first calls \texttt{sample\_distorted\_bounding\_box()} in \texttt{tf.image} to get the $(y_\mathrm{min}, y_\mathrm{max}, x_\mathrm{min}, x_\mathrm{max})$ bounding box for the crop. Omitting error handling and floating point arithmetic, calling \texttt{sample\_distorted\_bounding\_box()} with image height $h_o$, image width $w_o$, aspect ratio range $[r_\mathrm{min}, r_\mathrm{max}]$, and crop scale range $[a_\mathrm{min}, a_\mathrm{max}]$ follows Algorithm~\ref{alg:tf_crop}. While seemingly reasonable, sampling the crop height $h$ from a uniform distribution means that the cumulative distribution function (CDF) of the crop scale $a$ follows $\Pr(a < A) \propto \sqrt{A} - C$, so the probability density function (PDF) follows $f(A) \propto \frac{1}{\sqrt{A}}$. We can experimentally verify this distribution (\cref{fig:tf_area_distribution_approx}) and examine the probability heatmaps of the crop size for both landscape and portrait images (\cref{fig:tf_heat_map}). Note that the probability distribution is not equivariant to image transposition since the aspect ratio is sampled uniformly in linear space and sampled $h$ is clipped as $\min(\cdot, h_o)$ instead of rejection sampling.
\begin{algorithm}[t]
\begin{algorithmic}[1]
    \State {\bfseries Input:} $h_o, w_o, r_\mathrm{min}, r_\mathrm{max}, a_\mathrm{min}, a_\mathrm{max}$
    \For{$t=1$ {\bfseries to} \text{max\_attempts}}
    \State Sample $r \sim \mathcal{U}(r_\mathrm{min}, r_\mathrm{max})$
    \State $h_\mathrm{max} = \min(\sqrt{\frac{a_\mathrm{max}h_ow_o}{r}}, h_o)$
    \State $h_\mathrm{min} = \min(\sqrt{\frac{a_\mathrm{min}h_ow_o}{r}}, h_\mathrm{max})$
    \State Sample $h \sim \mathcal{U}(h_\mathrm{min}, h_\mathrm{max})$
    \State $w = hr$
    \State $a = hw$
    \If{$0 < h \le h_o$ and $0 < w \le w_o$ and $a_\mathrm{min}h_ow_o < a \le a_\mathrm{max}h_ow_o$}
    \State Sample $y \sim \mathcal{U}(0, h_o - h)$
    \State Sample $x \sim \mathcal{U}(0, w_o - w)$
    \State Return $(y, y + h, x, x + w)$
    \EndIf
    \EndFor
    \State Return $(0, h_o, 0, w_o)$ \Comment{Fallback to full image}
\end{algorithmic}
\caption{\texttt{sample\_distorted\_bounding\_box()}}
\label{alg:tf_crop}
\end{algorithm}
\subsection{PyTorch}
Inception crop is included in PyTorch \citep{Ansel_PyTorch_2_Faster_2024} as part of the TorchVision library \citep{TorchVision_maintainers_and_contributors_TorchVision_PyTorch_s_Computer_2016}, \texttt{RandomResizedCrop} (RRC, \citet{10.5555/3666122.3666890}) in \texttt{torchvision.transforms.v2}. Its counterpart to \texttt{sample\_distorted\_bounding\_box()} is the static method \texttt{get\_params()} which takes image tensor img, aspect ratio range $[r_\mathrm{min}, r_\mathrm{max}]$, and crop scale range $[a_\mathrm{min}, a_\mathrm{max}]$ as arguments and returns the bounding box as (top, left, $h$, $w$) following Algorithm~\ref{alg:torch_crop}. Unlike Algorithm~\ref{alg:tf_crop}, the resulting crop size distribution is indeed uniform and equivariant to image transposition (\cref{sec:torch_implementation}).
\begin{algorithm}[t]
\begin{algorithmic}[1]
    \State {\bfseries Input:} img, $r_\mathrm{min}, r_\mathrm{max}, a_\mathrm{min}, a_\mathrm{max}$
    \State $\_, h_o, w_o$ = \texttt{get\_dimensions(}img\texttt{)} 
    \For{$t=1$ {\bfseries to} 10}
        \State Sample $a \sim \mathcal{U}(a_\mathrm{min}h_ow_o, a_\mathrm{max}h_ow_o)$
        \State Sample $r_\mathrm{log} \sim \mathcal{U}(\log(r_\mathrm{min}), \log(r_\mathrm{max}))$
        \State $r = e^{r_\mathrm{log}}$ 
        \State $w = \sqrt{ar}$
        \State $h = \sqrt{\frac{a}{r}}$    
        \If{$0 < h \le h_o$ and $0 < w \le w_o$}
            \State Sample $y \sim \mathcal{U}(0, h_o - h)$
            \State Sample $x \sim \mathcal{U}(0, w_o - w)$
            \State Return $(y, x, h, w)$
        \EndIf
    \EndFor
    \State $r_o = \frac{w_o}{h_o}$
    \If{$r_o < r_\mathrm{min}$}
        \State $w = w_o$
        \State $h = \frac{w}{r_\mathrm{min}}$
    \ElsIf{$r_o > r_\mathrm{max}$}
        \State $h = h_o$
        \State $w = hr_\mathrm{max}$
    \Else
        \State $w = w_o$
        \State $h = h_o$
    \EndIf
    \State Return $(\frac{h_o - h}{2}, \frac{w_o - w}{2}, h, w)$ \Comment{Fallback to central crop}
\end{algorithmic}
\caption{\texttt{RandomResizedCrop.get\_params())}}
\label{alg:torch_crop}
\end{algorithm}

\section{Beta Crop}
\label{sec:beta}

\begin{figure}[t]
  \centering
  \includegraphics[width=0.4\linewidth]{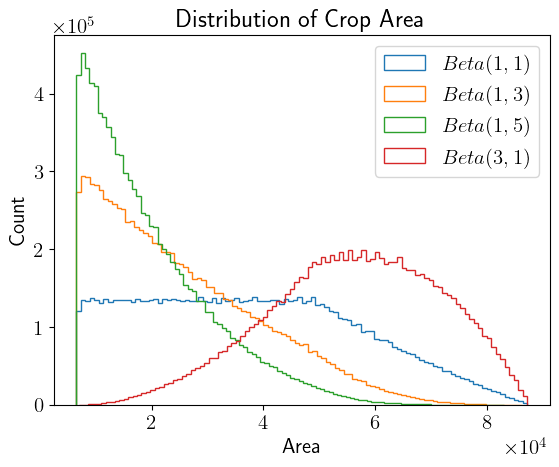}
  \caption{Crop area distribution of Beta crop with selected values of $\alpha$ and $\beta$ for landscape image ($h_o=256, w_o=512$), $r_{\min} = \frac{3}{4}, r_{\max} = \frac{4}{3}, a_{\min} = 0.05, a_{\max} = 1, N=10^7$ samples for each distribution. $\alpha = \beta = 1$ results in uniform distribution as expected except truncation at higher crop scale due to inadmissible aspect ratios.}
  \label{fig:beta_area_distribution}
\end{figure}

The TensorFlow implementation of Inception crop was \href{https://github.com/tensorflow/tensorflow/commit/f3a77378b4c056e76691c5eba350a022c11e00d4}{first committed 9 years ago} and has remained essentially unchanged. The fact that the resulting crop size discrepancy goes unnoticed motivates us to test its effect on model performance and the effect of crop scale distribution on model performance in general. Although we can generalize its practice of sampling the crop scale as $a = x^p, x \sim \mathcal{U}(a_{\min}^{\frac{1}{p}}, a_{\max}^{\frac{1}{p}})$ for $p$ other than $2$, this introduces a dependency of the distribution on the lower bound of the crop scale, which may not be desirable. We therefore decide to parameterize the the crop scale distribution with the beta distribution:
\begin{align*} 
a &= a_{\min} + x (a_{\max} - a_{\min}) \\
x &\sim Beta(\alpha, \beta)
\end{align*}

where the beta distribution $Beta(\alpha, \beta)$ is defined by the PDF 
$$
f(x; \alpha, \beta) = \frac{x^{\alpha-1}(1-x)^{\beta-1}}{\mathcal{B}(\alpha,\beta)}
$$

with $x \in [0, 1]$ and the beta function $\mathcal{B}(\alpha,\beta)$ as the normalization constant. In addition, we clip $a_{\max}$ to be the maximum possible crop scale when the image aspect ratio $r_o$ is out of the range $[r_{\min}, r_{\max}]$ (\cref{sec:beta_implementation}). When $\alpha = \beta = 1$, Beta crop reduces to Inception crop as a special case (\cref{fig:beta_area_distribution}). Although we later learn of prior experiments by \citet{NEURIPS2021_a2fe8c05} with Beta crop, we still find it worth exploring for 3 reasons:

\begin{enumerate}
  \item \citet{NEURIPS2021_a2fe8c05} only train ResNet-18 on ImageNet-1k with Beta crop, which is of a different class of model with about half as many parameters (12M) as ViT-S (22M).
  \item \citet{NEURIPS2021_a2fe8c05} train ResNet-18 with fixed training budget of 150 epochs but we have seen preliminary evidence that the optimal augmentation strength may depend on the training budget.
  \item Finally, \citet{NEURIPS2021_a2fe8c05} fix $\alpha = 1$ while varying $\beta \in \{0.1, 0.5, 1, 2, 3, 10\}$, which do not cover more modest deviation from uniform distribution such as the TensorFlow Inception crop's $f(A) \propto \frac{1}{\sqrt{A}}$ or $Beta(2, 1)$.
\end{enumerate}

\section{Experiments}
\label{sec:experiments}
We train a variant of ViT-S/16 described in \citet{beyer2022better} (sometimes called ``Simple ViT'', \citet{vit-pytorch}) on the ImageNet-1k dataset with PyTorch as the baseline, including the full setup of 1024 batch size, 10000 warm-up steps followed by cosine learning rate (LR) decay with AdamW \citep{loshchilov2018decoupled} optimizer, global average-pooling (GAP) \citep{raghu2021do}, 2D sin-cos positional encoding \citep{9711302} and Mixup \citep{zhang2018mixup} with $\alpha = 0.2$. We reimplement the RandAugment \citep{NEURIPS2020_d85b63ef} in the Big Vision codebase \citep{big_vision} except the broken \texttt{Contrast} transform, which we elect to fix (\cref{sec:randaugment_implementation}). For convenience and consistency, we also reimplement the TensorFlow Inception crop almost line-by-line in PyTorch and verify that the PyTorch reimplementation results in indistinguishable crop size distributions from that of the original (\cref{sec:torch_reimplementation}). For both PyTorch Inception crop and reimplemented TensorFlow Inception crop, we train the models for $\{30, 60, 90, 150, 300\}$ epochs with crop scale lower bound $a_{\min} \in \{0.025, 0.05, 0.1, 0.15, 0.2, 0.25\}$. We then sweep the the parameter space of Beta crop in three disjoint sets:

\begin{enumerate}
  \item Positive skew: $a_{\min} \in \{0.25, 0.33, 0.5\}, \alpha=1, \beta \in \{3, 5\}$ that compensates large lower bound of crop scale with more samples near the lower bound;
  \item Negative skew: $a_{\min} \in \{0.05, 0.1\}, \alpha \in \{2, 3, 5\}, \beta = 1$ that compensates small lower bound of crop scale with less samples near the lower bound.
  \item Matching strength: $a_{\min} = 0, \alpha \in \{1.149, 1.287, 1.551, 1.818, 2.099, 2.403\}, \beta = 1$ that are designed to match the augmentation strength of PyTorch Inception crop with $a_{\min} \in \{0.025, 0.05, 0.1, 0.15, 0.2, 0.25\}$.
\end{enumerate}

Finally, we replicate some of the results with constrained Scion optimizer \citep{pethick2025training}. We adopt the model modifications, batch size 4096, cosine LR decay with no warm-up, and maximum LR for DeiT-base \citep{pmlr-v139-touvron21a} from \citet{pethick2025training} and verify that the maximum LR remains optimal for the 90ep experiment by doubling and halving it. We then perform our own weight decay (WD) sweep through 4 values for all training budgets (\cref{sec:scion_setup}). With the resulting setup we compare the model performance with PyTorch Inception crop and fixed $a_{\min} = 0.05$ vs. the optimized $a_{\min}$ for each training budget, identified in the AdamW experiments. For each setting we train $N=3$ models with different random seeds. We evaluate the models on the validation set with both the original labels and the multi-label ReaL \citep{beyer2020we}. The models are trained on either $8 \times$A100 or $2 \times$H100 instances depending on the availability, but with standardized Python and package versions (Python 3.12.x, torch==2.5.1+cu124, and torchvision==0.20.1+cu124).
\section{Results}
\label{sec:results}
\subsection{Inception crop implementations}
\label{sec:inception_crop_results}

\begin{figure*}[hbt!]
    \begin{subfigure}{0.33\textwidth}
    \centering
    \includegraphics[width=\textwidth]{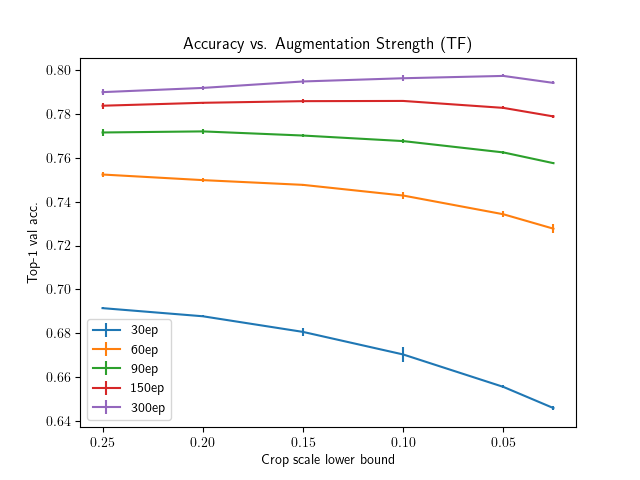}%
    \caption*{TensorFlow}
    \end{subfigure}%
    \begin{subfigure}{0.33\textwidth}
    \centering
    \includegraphics[width=\textwidth]{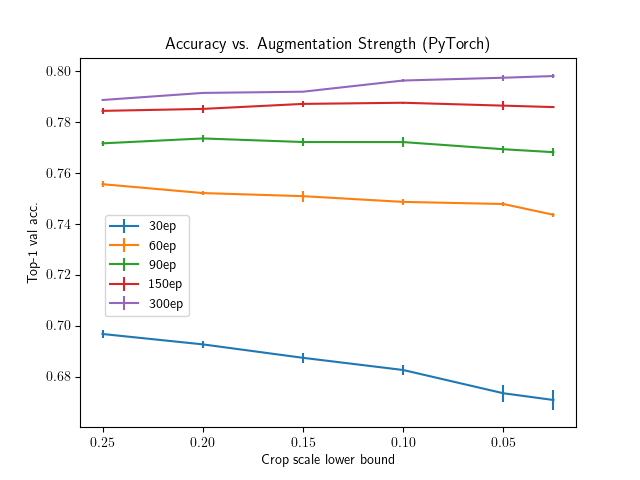}%
    \caption*{PyTorch}
    \end{subfigure}%
    \begin{subfigure}{0.33\textwidth}
    \centering
    \includegraphics[width=\textwidth]{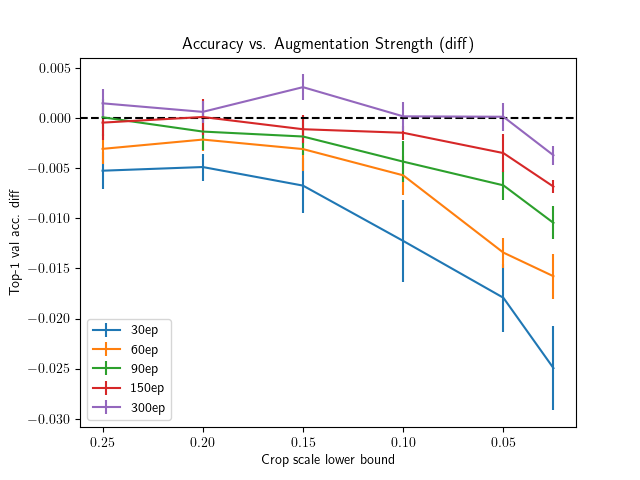}%
    \caption*{$\Delta$(TensorFlow - PyTorch)}
    \end{subfigure}%
    \caption{Top-1 val. accuracy (original label), TensorFlow vs. PyTorch Inception crop, with sample standard deviation $\sigma_\mathrm{tf}, \sigma_\mathrm{torch}, (\sigma^2_\mathrm{tf} + \sigma^2_\mathrm{torch})^{\frac{1}{2}}$ plotted as error bar. As the training budget increases from 30 epochs to 300 epochs, the optimal crop scale lower bound $a_\mathrm{min}$ decreases (stronger augmentation).}
    \label{fig:acc_vs_aug}
\end{figure*}

\begin{figure*}[hbt!]
    \begin{subfigure}{0.33\textwidth}
    \centering
    \includegraphics[width=\textwidth]{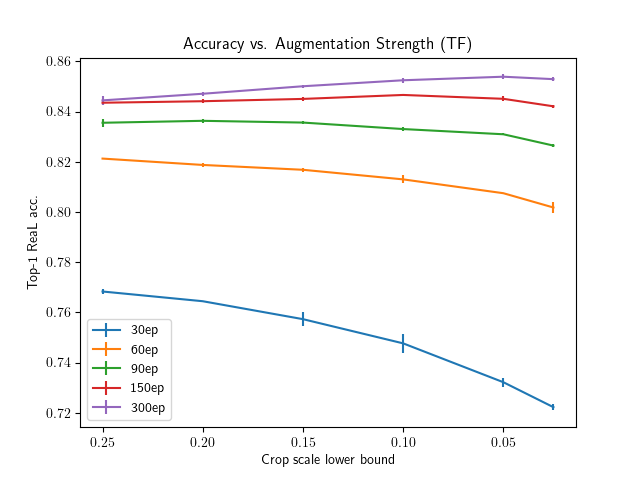}
    \caption*{TensorFlow}
    \end{subfigure}%
    \begin{subfigure}{0.33\textwidth}
    \centering
    \includegraphics[width=\textwidth]{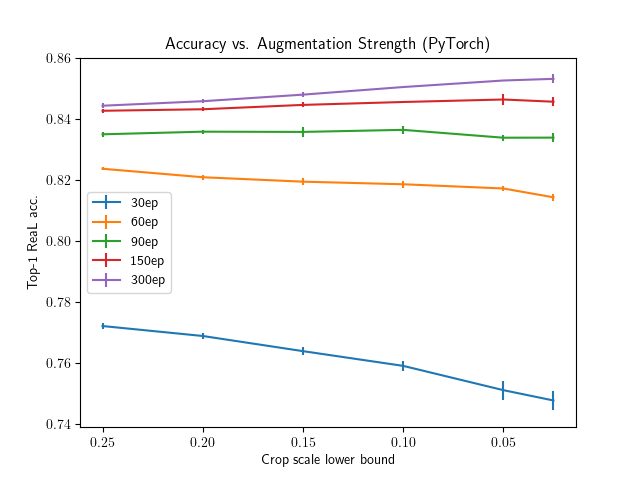}
    \caption*{PyTorch}
    \end{subfigure}%
    \begin{subfigure}{0.33\textwidth}
    \centering
    \includegraphics[width=\textwidth]{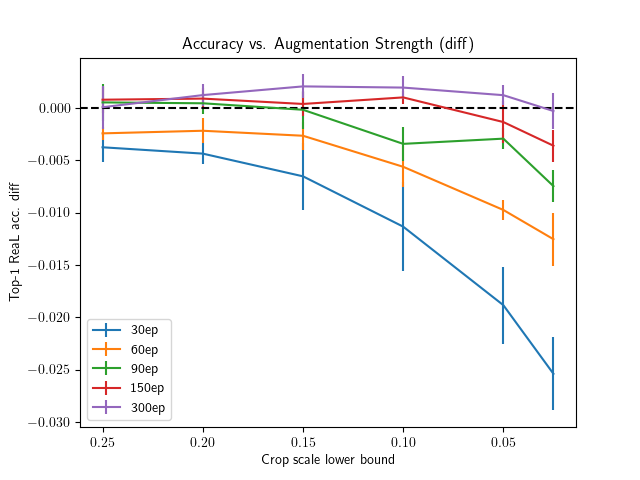}
    \caption*{$\Delta$(TensorFlow - PyTorch)}
    \end{subfigure}%
    \caption{Top-1 val. accuracy (ReaL), TensorFlow vs. PyTorch Inception crop, with sample standard deviation $\sigma_\mathrm{tf}, \sigma_\mathrm{torch}, (\sigma^2_\mathrm{tf} + \sigma^2_\mathrm{torch})^{\frac{1}{2}}$ plotted as error bar. As the training budget increases from 30 epochs to 300 epochs, the optimal crop scale lower bound $a_\mathrm{min}$ decreases (stronger augmentation).}
    \label{fig:acc_vs_aug_real}
\end{figure*}

Here are the results of training ViT-S/16 models on ImageNet-1k with implementations of Inception crop, varying training budget and $a_\mathrm{min}$ (\cref{table:tf_vs_torch_original,fig:acc_vs_aug,sec:inception_full}). The optimal crop scale lower bound $a_\mathrm{min}$ shifts from $\ge 0.25$ to $\le 0.05$ as we increase the training budget from 30 epochs to 300 epochs. Compared to the PyTorch counterpart, the (reimplemented) TensorFlow Inception crop is almost never favorable and becomes worse with smaller $a_\mathrm{min}$, likely due to the increasing positive skew of the crop scale distribution. Evaluation using the multi-label ReaL shifts the top-1 val. accuracy by +6--7\% but the results stand (\cref{table:tf_vs_torch_real,fig:acc_vs_aug_real,sec:inception_full}). The only settings in which the TensorFlow Inception corp performs better are with 150--300ep training budget and $a_\mathrm{min} \ge 0.05$, which can be explained by the differences in augmentation strength (\cref{sec:inception_equal_strength_comparison}).

\begin{table*}[hbt!]
\small
\centering
\begin{tabular}{|c||c|c|c|c|c|c|c|c|c|c|c|c|}
\hline
& \multicolumn{6}{c|}{TensorFlow Inception crop} & \multicolumn{6}{c|}{PyTorch Inception crop} \\
\hline \hline
$a_\mathrm{min}$ & 0.025 & 0.05 & 0.1 & 0.15 & 0.2 & 0.25 & 0.025 & 0.05 & 0.1 & 0.15 & 0.2 & 0.25 \\
\hline
30ep & 64.59 & 65.55 & 67.03 & 68.06 & 68.78 & \textbf{69.15} & 67.08 & 67.35 & 68.26 & 68.74 & 69.26 & \textbf{69.67} \\
\hline
60ep & 72.77 & 73.43 & 74.29 & 74.77 & 74.99 & \textbf{75.25} & 74.35 & 74.77 & 74.86 & 75.08 & 75.20 & \textbf{75.55} \\
\hline
90ep & 75.76 & 76.25 & 76.77 & 77.03 & \textbf{77.21} & 77.16 & 76.81 & 76.92 & 77.21 & 77.21 & \textbf{77.35} & 77.16 \\
\hline
150ep & 77.90 & 78.29 & \textbf{78.60} & 78.59 & 78.52 & 78.39 & 78.58 & 78.64 & \textbf{78.75} & 78.70 & 78.51 & 78.43 \\
\hline
300ep & 79.43 & \textbf{79.75} & 79.64 & 79.49 & 79.20 & 79.01 & \textbf{79.80} & 79.73 & 79.62 & 79.18 & 79.13 & 78.86 \\
\hline
\end{tabular}
\caption{Mean top-1 val. accuracy (original label), TensorFlow vs. PyTorch Inception crop. As the training budget increases from 30 epochs to 300 epochs, the optimal crop scale lower bound $a_\mathrm{min}$ decreases (stronger augmentation).}
\label{table:tf_vs_torch_original}
\end{table*}

\begin{table*}[hbt!]
\small
\centering
\begin{tabular}{|c||c|c|c|c|c|c|c|c|c|c|c|c|}
\hline
& \multicolumn{6}{c|}{TensorFlow Inception crop} & \multicolumn{6}{c|}{PyTorch Inception crop} \\
\hline \hline
$a_\mathrm{min}$ & 0.025 & 0.05 & 0.1 & 0.15 & 0.2 & 0.25 & 0.025 & 0.05 & 0.1 & 0.15 & 0.2 & 0.25 \\
\hline
30ep & 72.23 & 73.22 & 74.77 & 75.73 & 76.45 & \textbf{76.84} & 74.77 & 75.11 & 75.90 & 76.39 & 76.88 & \textbf{77.21} \\
\hline
60ep & 80.18 & 80.75 & 81.30 & 81.68 & 81.88 & \textbf{82.13} & 81.44 & 81.73 & 81.86 & 81.95 & 82.09 & \textbf{82.37} \\
\hline
90ep & 82.65 & 83.10 & 83.31 & 83.57 & \textbf{83.64} & 83.56 & 83.39 & 83.39 & \textbf{83.65} & 83.58 & 83.59 & 83.50 \\
\hline
150ep & 84.21 & 84.51 & \textbf{84.67} & 84.51 & 84.42 & 84.36 & 84.57 & \textbf{84.65} & 84.56 & 84.47 & 84.33 & 84.28 \\
\hline
300ep & 85.30 & \textbf{85.39} & 85.25 & 85.01 & 84.71 & 84.45 & \textbf{85.32} & 85.27 & 85.06 & 84.81 & 84.59 & 84.44 \\
\hline
\end{tabular}
\caption{Mean top-1 val. accuracy (ReaL), TensorFlow vs. PyTorch Inception crop. As the training budget increases from 30 epochs to 300 epochs, the optimal crop scale lower bound $a_\mathrm{min}$ decreases (stronger augmentation).}
\label{table:tf_vs_torch_real}
\end{table*}

\begin{figure*}
\begin{minipage}{0.5\textwidth}
    \raggedleft
    \small
    \begin{tabular}{|c||c||c|c|c|}
        \hline
         & $a_\mathrm{min}$ & 0.25 & 0.33 & 0.5 \\
        \hline
        \multirow{5}{*}{$\beta = 3$} & 30ep & 68.38$\pm 0.08$ & 69.32$\pm 0.20$ & \textbf{70.35}$\pm 0.24$ \\
        \cline{2-5}
         & 60ep & 74.90$\pm 0.10$ & 75.17$\pm 0.07$ & \textbf{75.42}$\pm 0.02$ \\
        \cline{2-5}
         & 90ep & 76.69$\pm 0.11$ & \textbf{76.93}$\pm 0.26$ & 76.70$\pm 0.12$ \\
        \cline{2-5}
         & 150ep & 78.14$\pm 0.19$ & \textbf{78.18}$\pm 0.14$ & 77.57$\pm 0.15$ \\
        \cline{2-5}
         & 300ep & \textbf{78.85}$\pm 0.14$ & 78.48$\pm 0.07$ & 77.63$\pm 0.03$ \\
        \hline
        \multirow{5}{*}{$\beta = 5$} & 30ep & 67.31$\pm 0.20$ & 68.92$\pm 0.24$ & \textbf{70.24}$\pm 0.21$ \\
        \cline{2-5}
         & 60ep & 74.01$\pm 0.05$ & 74.81$\pm 0.18$ & \textbf{75.42}$\pm 0.20$ \\
        \cline{2-5}
         & 90ep & 76.31$\pm 0.10$ & 76.69$\pm 0.05$ & \textbf{76.71}$\pm 0.18$ \\
        \cline{2-5}
         & 150ep & 77.86$\pm 0.04$ & \textbf{77.90}$\pm 0.03$ & 77.48$\pm 0.07$ \\
        \cline{2-5}
         & 300ep & \textbf{78.61}$\pm 0.01$ & 78.42$\pm 0.15$ & 77.59$\pm 0.14$ \\
        \hline
    \end{tabular}
\end{minipage}%
\hfill
\begin{minipage}{0.33\textwidth}
    \raggedright
    \includegraphics[width=\textwidth]{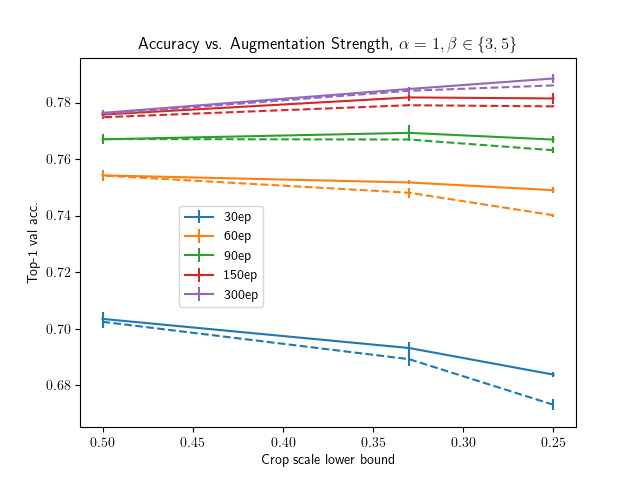}
\end{minipage}
\caption{Top-1 val. accuracy (original label), Beta crop with $\alpha=1$. Solid line: $\beta=3$, Dashed line: $\beta=5$}
\label{fig:beta_val_full}
\end{figure*}

\begin{figure*}
\centering
\begin{minipage}{0.5\textwidth}
    \small
    \centering
    \begin{tabular}{|c||c||c|c|}
    \hline
     & $a_\mathrm{min}$ & 0.05 & 0.1 \\
    \hline
    \multirow{5}{*}{$\alpha = 2$} & 30ep & 69.68$\pm 0.04$ & \textbf{70.04}$\pm 0.27$ \\
    \cline{2-4}
     & 60ep & 75.46$\pm 0.22$ & \textbf{75.66}$\pm 0.15$ \\
    \cline{2-4}
     & 90ep & \textbf{77.35}$\pm 0.24$ & 77.29$\pm 0.14$ \\
    \cline{2-4}
     & 150ep & \textbf{78.56}$\pm 0.11$ & 78.28$\pm 0.17$ \\
    \cline{2-4}
     & 300ep & \textbf{79.07}$\pm 0.24$ & 78.85$\pm 0.01$ \\
    \hline
    \multirow{5}{*}{$\alpha = 3$} & 30ep & 70.24$\pm 0.14$ & \textbf{70.28}$\pm 0.17$ \\
    \cline{2-4}
     & 60ep & \textbf{75.58}$\pm 0.21$ & 75.55$\pm 0.08$ \\
    \cline{2-4}
     & 90ep & 77.11$\pm 0.07$ & \textbf{77.12}$\pm 0.16$ \\
    \cline{2-4}
     & 150ep & \textbf{78.15}$\pm 0.10$ & 78.12$\pm 0.21$ \\
    \cline{2-4}
     & 300ep & \textbf{78.44}$\pm 0.12$ & 78.23$\pm 0.12$ \\
    \hline
    \multirow{5}{*}{$\alpha = 5$} & 30ep & 70.50$\pm 0.12$ & \textbf{70.66}$\pm 0.20$ \\
    \cline{2-4}
     & 60ep & \textbf{75.34}$\pm 0.29$ & 75.31$\pm 0.08$ \\
    \cline{2-4}
     & 90ep & \textbf{76.80}$\pm 0.11$ & 76.66$\pm 0.15$ \\
    \cline{2-4}
     & 150ep & \textbf{77.43}$\pm 0.11$ & 77.27$\pm 0.16$ \\
    \cline{2-4}
     & 300ep & \textbf{77.60}$\pm 0.17$ & 77.44$\pm 0.13$ \\
    \hline
    \end{tabular}
\end{minipage}%
\begin{minipage}{0.5\textwidth}
    \centering
    \includegraphics[width=\linewidth]{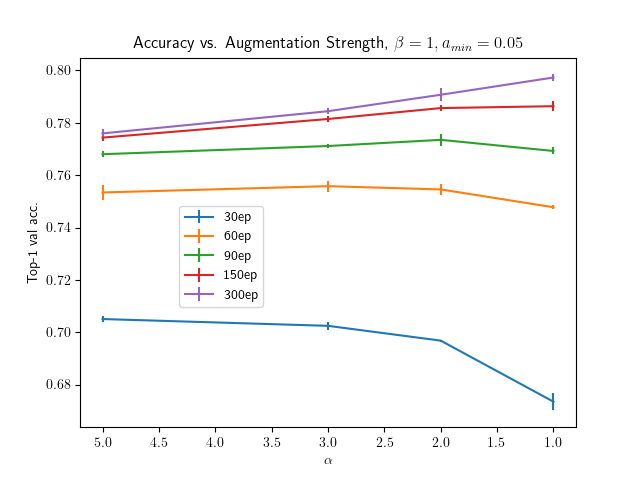}
\end{minipage}
\caption{Top-1 val. accuracy (original label), Beta crop with $\beta=1$. Top-1 val. accuracy vs. $\alpha$, $a_\mathrm{min} = 0.05$ is plotted on the right. For $\alpha=\beta=1$, we reuse the data points of the PyTorch Inception crop experiments.}
\label{fig:alpha_val_full}
\end{figure*}

\subsection{Beta crop}
\label{sec:beta_crop_results}
Among the Beta crop experiments, we find that higher positive skew of the distribution does not compensate for higher $a_\mathrm{min}$ (\cref{fig:beta_val_full}). That is, none of the set of experiments with $a_\mathrm{min} = 0.25, \alpha=1, \beta \in \{3, 5\}$ improves upon the corresponding experiment with the PyTorch Inception crop with the same $a_\mathrm{min}$. This suggests that even more samples with smaller crops in the distribution do not help the model generalize: The model needs to see out-of-distribution crops with lower crop scale.

On the other hand, negative skew of the distribution does compensate for lower $a_\mathrm{min}$ (\cref{fig:alpha_val_full}). That is, the set of experiments with $a_\mathrm{min} \in \{0.05, 0.1\}, \alpha \in \{2, 3, 5\}, \beta = 1$ does improve upon the corresponding experiment with the PyTorch Inception crop with the same $a_\mathrm{min}$ with training budgets 30--90 epochs. Quantitatively just like PyTorch Inception crop with $a_\mathrm{min} = 0.05$ which crops 5.3\% of the samples with crop scale between $0.05$ and $0.1$, Beta crop with $a_\mathrm{min} = 0.05, \alpha=2, \beta=1$ crops 5.3\% of the samples below crop scale $0.268$. With the working hypothesis that the smallest 5.3\% of the crops determine the model performance, we expect it to be comparable to PyTorch Inception crop with $a_\mathrm{min} = 0.227$ that also crops 5.3\% of the samples below crop scale $0.268$. Reexamining \cref{fig:alpha_val_full} with this insight, we can see that the former indeed has performance between that of the PyTorch Inception crop with $a_\mathrm{min} = 0.2$ and $0.25$ for all training budgets 30--300ep (\cref{table:tf_vs_torch_original}). This observation motivates the matching strength experiments below.

\begin{figure*}
    \begin{minipage}{0.5\textwidth}
    \centering
    \includegraphics[width=0.8\textwidth]{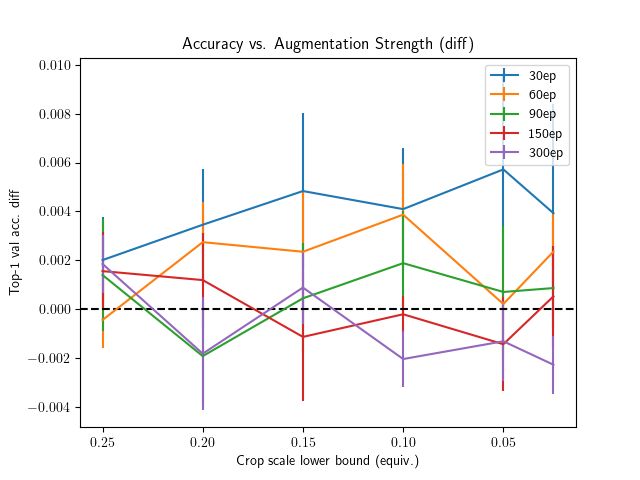}
    \end{minipage}%
    \begin{minipage}{0.5\textwidth}
    \centering
    \includegraphics[width=0.8\textwidth]{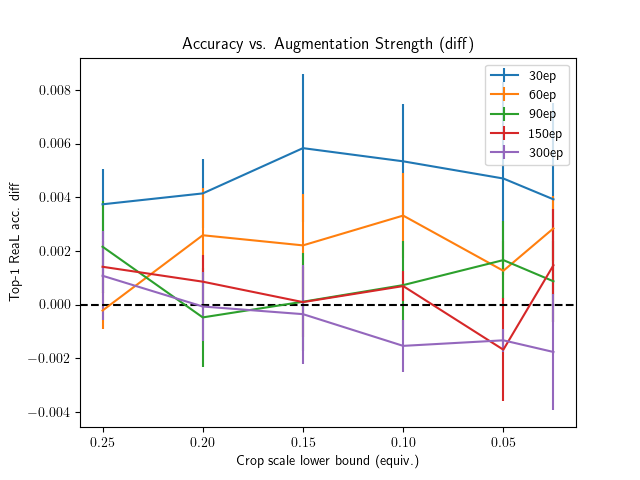}
    \end{minipage}%
    \caption{Diff. in top-1 val. accuracy (Beta crop, matching strength - PyTorch), original label (left) and ReaL (right), with $(\sigma^2_\mathrm{beta} + \sigma^2_\mathrm{torch})^{\frac{1}{2}}$ plotted as error bar.}
  \label{fig:matching_strength_diff}
\end{figure*}

Beta crop with $a_{\min} = 0, \alpha \in \{1.149, 1.287, 1.551, 1.818, 2.099, 2.403\}, \beta = 1$ have the same 5th percentile crop scale as that of PyTorch Inception crop with $a_{\min} \in \{0.025, 0.05, 0.1, 0.15, 0.2, 0.25\}$, respectively. \Eg, $\Pr(a < 0.07375) = 0.05$ with $a \sim Beta(\alpha=1.149, \beta=1)$ since $a_{\min} + 0.05(1 - a_{\min}) = 0.07375$ when $a_{\min} = 0.025$. Models trained in these experiment match or exceed the performance of the PyTorch Inception crop in most settings. In particular, softer cutoff of Beta crop allows it to compromise less of the performance even when the augmentation strength is not optimally tuned for the given training budget (\cref{table:matching_strength,fig:matching_strength_diff}).

\begin{table*}[hbt!]
\small
\centering
\begin{tabular}{|c||c|c|c|c|c|c|c|c|c|c|c|c|}
\hline
& \multicolumn{6}{c|}{Original label} & \multicolumn{6}{c|}{ReaL} \\
\hline \hline
$\alpha$ & 1.149 & 1.287 & 1.551 & 1.818 & 2.099 & 2.403 & 1.149 & 1.287 & 1.551 & 1.818 & 2.099 & 2.403  \\
\hline
30ep & 67.47 & 67.92 & 68.67 & 69.22 & 69.61 & \textbf{69.87} & 75.16 & 75.58 & 76.44 & 76.97 & 77.30 & \textbf{77.58}  \\
\hline
60ep & 74.59 & 74.79 & 75.24 & 75.32 & 75.48 & \textbf{75.51} & 81.72 & 81.85 & 82.20 & 82.17 & \textbf{82.35} & 82.35  \\
\hline
90ep & 76.89 & 76.99 & \textbf{77.39} & 77.25 & 77.15 & 77.29 & 83.48 & 83.56 & \textbf{83.72} & 83.59 & 83.54 & 83.72  \\
\hline
150ep & 78.63 & 78.49 & \textbf{78.73} & 78.59 & 78.63 & 78.59 & \textbf{84.72} & 84.48 & 84.63 & 84.48 & 84.41 & 84.42  \\
\hline
300ep & 79.57 & \textbf{79.60} & 79.42 & 79.27 & 78.95 & 79.04 & \textbf{85.15} & 85.14 & 84.90 & 84.77 & 84.58 & 84.55  \\
\hline
\end{tabular}
\caption{Mean top-1 val. accuracy of Beta crop, matching strength, original label (left) and ReaL (right). As the training budget increases from 30 epochs to 300 epochs, the optimal $\alpha$ decreases (stronger augmentation). $\alpha = 1.149$ may be an exception possibly due to too many crops near zero area. In future iterations we may want small but nonzero $a_{\min}$ (say $0 < a_{\min} \le 0.025$) instead while keeping the same 5th percentile crop scale.}
\label{table:matching_strength}
\end{table*}

For all of the Beta crop experiments the correlation between the optimal augmentation strength and training budget continues to hold, and evaluating the models using the multi-label ReaL does not change the results (\cref{sec:beta_full}).

\subsection{Saliency sparsity vs. training budget}
\label{sec:saliency_sparsity}

\begin{figure*}
    \begin{minipage}{0.5\textwidth}
    \centering
    \includegraphics[width=0.8\textwidth]{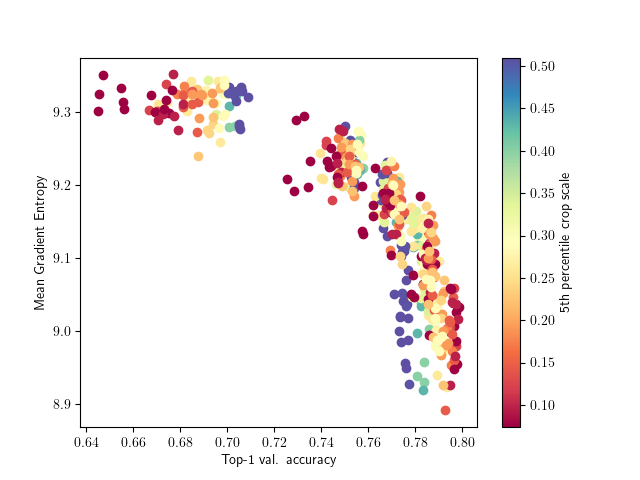}
    \end{minipage}%
    \begin{minipage}{0.5\textwidth}
    \centering
    \includegraphics[width=0.8\textwidth]{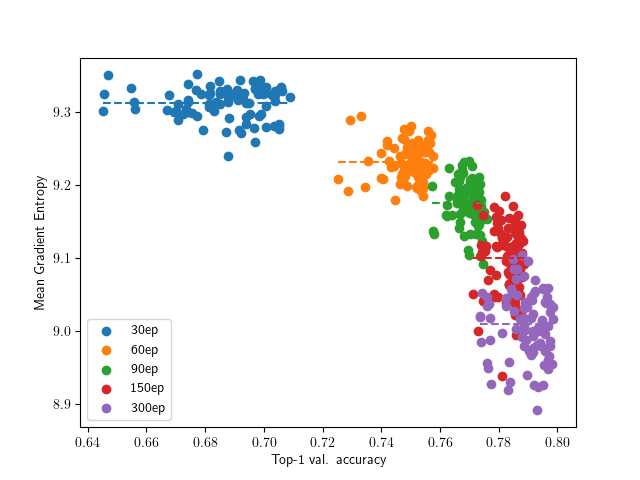}
    \end{minipage}%
    \caption{Mean gradient entropy $\overline{S_\mathrm{grad}}$ defined in \cref{eq:grad_entropy} vs. top-1 val. accuracy for all 450 Inception crop and Beta crop experiments with AdamW optimizer, color-coded by the 5th percentile crop scale (left) and training budget (right). For reference and sanity check, $S_\mathrm{grad} \le \log(224^2) = 10.82$ given our input resolution. Dashed lines mark the average of the experiments of the given training budget.}
    \label{fig:grad_entropy_scatter}
\end{figure*}

Since the crop scale determines the proportion of the image on which the model is trained, we initially hypothesize that crop scale distribution may affect saliency sparsity of the model. We first measure it using the rectified gradient of the class logit on the pixels as the proxy. That is, given the logit $y^c$ of the target class $c$ returned by the model, we compute its gradient on the pixels $a_\mathrm{ijk}$, summed over the RGB channels $k \in [0\ldotp \ldotp2]$ and clipped as $\max(0, \cdot)$:

\begin{equation}
g_\mathrm{ij} = \max(0, \sum\limits_\mathrm{k} \frac{\partial y^c}{\partial a_\mathrm{ijk}})
\end{equation}

where $i \in [0\ldotp \ldotp h-1]$ and $j \in [0\ldotp \ldotp w-1]$, $h=w=224$ for our input resolution. We can then measure the entropy of the distribution of the rectified gradient after normalization:

\begin{equation} \label{eq:grad_entropy}
S_\mathrm{grad} = - \sum\limits_\mathrm{ij} g'_\mathrm{ij} \log g'_\mathrm{ij}\,, \quad g'_\mathrm{ij} = \frac{g_\mathrm{ij}}{\sum\limits_\mathrm{ij} g_\mathrm{ij}}
\end{equation}

\begin{figure*}
    \begin{minipage}{0.5\textwidth}
    \centering
    \includegraphics[width=0.8\textwidth]{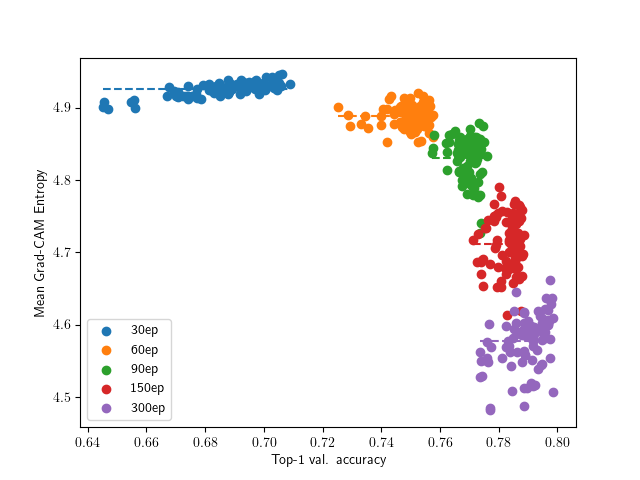}
    \end{minipage}%
    \begin{minipage}{0.5\textwidth}
    \centering
    \includegraphics[width=0.8\textwidth]{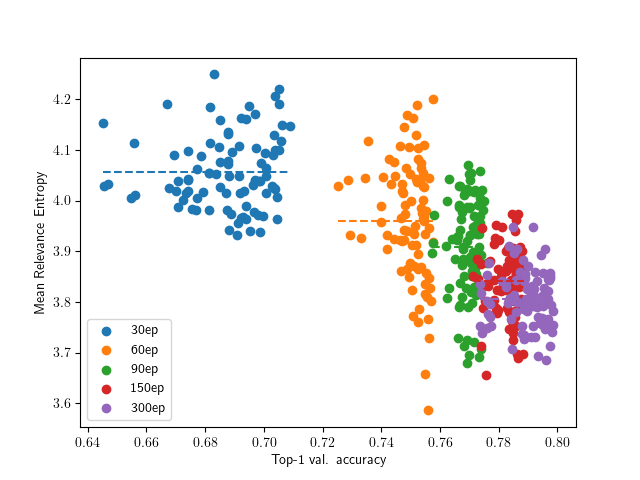}
    \end{minipage}%
    \caption{Mean Grad-CAM entropy $\overline{S_\mathrm{grad\text{-}cam}}$ (left) and mean relevance entropy $\overline{S_\mathrm{rel}}$ (right) vs. top-1 val. accuracy for all 450 Inception crop and Beta crop experiments with AdamW optimizer, color-coded by the training budget. For reference and sanity check, $S_\mathrm{grad\text{-}cam}, S_\mathrm{rel} \le \log(14^2) = 5.28$ given our input resolution and patch size. Dashed lines mark the average of the experiments of the given training budget.}
  \label{fig:rel_entropy_vs_acc}
\end{figure*}

Using the 50000 images and target class of the ImageNet-1k validation, we find no correlation between the mean gradient entropy $\overline{S_\mathrm{grad}}$ and the crop scale distribution with which the model is trained. Instead, it correlates with the training budget (\cref{fig:grad_entropy_scatter}). If we inspect each individual model, we can see that the distribution of $S_\mathrm{grad}$ for the 50000 validation set images shifts to the left as we increase the training budget (\cref{sec:grad_entropy_plots}). We repeat the measurement with Grad-CAM \citep{selvaraju2017grad} and LRP of the tokens of the last layer (\citet{binder2016layer}, \cref{sec:rel_entropy_plots}) and compute the entropy of their distribution, $S_\mathrm{grad\text{-}cam} = - \sum\limits_\mathrm{ij} L^c_\mathrm{ij} \log L^c_\mathrm{ij}$ and $S_\mathrm{rel} = - \sum\limits_\mathrm{ij} R'_\mathrm{ij} \log R'_\mathrm{ij}$, $i,j \in [0\ldotp \ldotp 13]$, after normalization. We find that the result stands: The entropy of saliency always decreases as we increase the training budget (\cref{fig:rel_entropy_vs_acc}).

\subsection{Partial replication with Scion optimizer}
\label{sec:scion_rep}
\begin{figure*}[hbt!]
    \begin{subfigure}{0.33\textwidth}
    \centering
    \includegraphics[width=\textwidth]{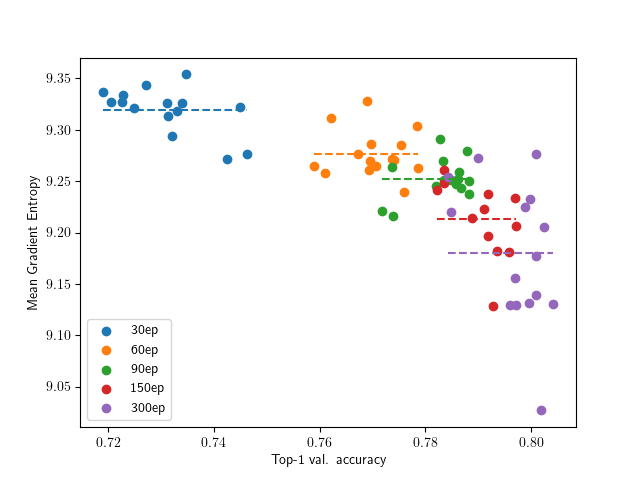}
    \caption*{$\overline{S_\mathrm{grad}}$}
    \end{subfigure}%
    \begin{subfigure}{0.33\textwidth}
    \centering
    \includegraphics[width=\textwidth]{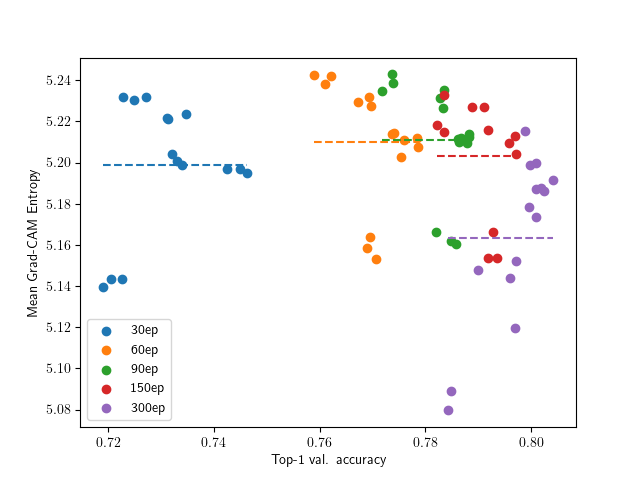}
    \caption*{$\overline{S_\mathrm{grad\text{-}cam}}$}
    \end{subfigure}%
    \begin{subfigure}{0.33\textwidth}
    \centering
    \includegraphics[width=\textwidth]{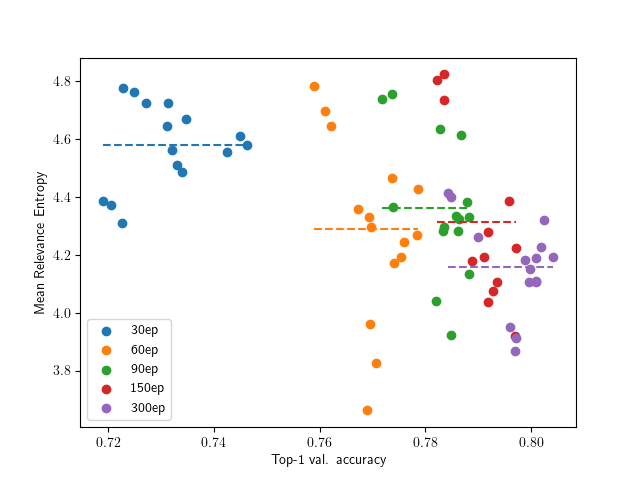}
    \caption*{$\overline{S_\mathrm{rel}}$}
    \end{subfigure}%
    \caption{Mean gradient entropy $\overline{S_\mathrm{grad}}$ (left), mean Grad-CAM entropy $\overline{S_\mathrm{grad\text{-}cam}}$ (middle), and mean relevance entropy $\overline{S_\mathrm{rel}}$ (right) vs. top-1 val. accuracy for all 72 Scion experiments including that of \cref{table:scion_exp} and WD sweep (\cref{table:scion_wd}), color-coded by training budget.}
    \label{fig:scion_entropy_vs_acc}
\end{figure*}

Finally, we replicate the spontaneous sparsification of saliency and the correlation between optimal augmentation strength and training budget with constrained Scion \citep{pethick2025training} optimizer instead of AdamW. As expected from \cref{table:tf_vs_torch_real}, PyTorch Inception crop with $a_\mathrm{min} = 0.25, 0.25, 0.1, 0.025$ outperform $a_\mathrm{min} = 0.05$ for training budgets 30ep, 60ep, 90ep, and 300ep respectively (\cref{table:scion_exp}). The entropy of saliency also generally decreases as we increase the training budget for the Scion experiments (\cref{fig:scion_entropy_vs_acc}).

\begin{table*}[hbt!]
\small
\centering
\begin{tabular}{|c||c|c|c|c|c|c|c|c|c|c|c|c|}
\hline
& \multicolumn{2}{c|}{Original label} & \multicolumn{2}{c|}{ReaL} \\
\hline \hline
& baseline & optimized & baseline & optimized \\
\hline
30ep & 73.31$\pm 0.09$ & \textbf{74.46}$\pm 0.19$ & 80.83$\pm 0.07$ & \textbf{81.70}$\pm 0.08$ \\
\hline
60ep & 77.44$\pm 0.09$ & \textbf{77.76}$\pm 0.15$ & 84.09$\pm 0.10$ & \textbf{84.16}$\pm 0.16$ \\
\hline
90ep & 78.68$\pm 0.09$ & \textbf{78.78}$\pm 0.09$ & 84.94$\pm 0.05$ & \textbf{84.96}$\pm 0.07$ \\
\hline
150ep & 79.65$\pm 0.07$ & --- & 85.41$\pm 0.05$ & --- \\
\hline
300ep & 80.10$\pm 0.14$ & \textbf{80.23}$\pm 0.17$ & 85.66$\pm 0.21$ & \textbf{85.71}$\pm 0.15$ \\
\hline
\end{tabular}
\caption{Top-1 val. accuracy of Scion experiments, $a_\mathrm{min} = 0.05$ baseline vs. optimized $a_\mathrm{min}$ for the training budget, original label (left) and ReaL (right). As the training budget increases from 30 epochs to 300 epochs, the optimal crop scale lower bound $a_\mathrm{min}$ decreases (stronger augmentation) while $a_\mathrm{min} = 0.05$ is already expected to be optimal for 150ep.}
\label{table:scion_exp}
\end{table*}

\section{Conclusions}
\label{sec:conclusions}
In conclusion, we find that 

\begin{enumerate}
  \item Optimal augmentation strength depends on the training budget. Higher training budget requires stronger data augmentation (\cref{sec:inception_crop_results}).
  \item Lower tail of the distribution of the crop scale determines the augmentation strength of Inception crop (\cref{sec:beta_crop_results}).
  \item Models trained with higher training budget exhibit sparser saliency, regardless of the augmentation strength (\cref{sec:saliency_sparsity,sec:scion_rep}) or weight decay (\cref{sec:scion_setup}).
\end{enumerate}

In practice, 1. can further complicate efforts to make training budget more flexible \citep{hagele2024scaling} or schedule-free \citep{defazio2024the} for deep vision models. Generalized curriculum learning \citep{10530470} though may be a viable method of mitigating these complications. Another research direction that may be worth pursuing is whether 1. changes the trade-off between model size and number of training epochs. \Eg, it may be advantageous to opt for training a larger model for fewer epochs with weaker augmentation, especially if we want to alleviate the negative effects of strong data augmentation due to semantic degradation \citep{10.5555/3666122.3666890}. Our partial mitigation in light of 2. is to propose Beta crop with matching strength whose softer cutoff
allows it to optimize model performance across training budgets with less compromise (\cref{table:matching_strength,fig:matching_strength_diff}). The TensorFlow implementation of the Inception crop is (likely unintentionally) the opposite with positive skew that increases with lower $a_\mathrm{min}$ and consequently underperforms in most settings (\cref{table:tf_vs_torch_original,table:tf_vs_torch_real}) even if we adopt a unified augmentation strength measure based on 2. (\cref{sec:inception_equal_strength_comparison}). We therefore recommend the TensorFlow / JAX community to fix or replace their implementation and urge caution while comparing results between the TensorFlow / JAX vs. the PyTorch ecosystems for models trained with ``Inception crop''.

From the theoretical perspective, it is not clear why 1. and 3. are true. One possibility for 1. is that stronger augmentation helps delay memorization of limited training examples, which would otherwise prevent further generalization \citep{bayat2025the}. Assumed that conversely delaying memorization past the end of the training budget leads to underfitting, the optimal augmentation needs to be tuned such that memorization happens at the end, leading to our observation.
\section{Limitations}
\label{sec:limitations}
The main limitations of our study are due to our focus on training ViT-S/16 models on the ImageNet-1k dataset, so it leaves the question open whether the results generalize to other model architectures, datasets, or larger/smaller models. However, it is worth pointing out again that \citet{steiner2022how} show in Figure 4 that AugReg becomes helpful for training a wide variety of ViT and ResNet models on ImageNet-21k as we increase the training budget from 30ep to 300ep, thus lending support to conclusion 1. Similarly, \citet{NEURIPS2021_a2fe8c05} report in Section 2.2 that Beta crop with $\alpha=1, \beta \in \{0.1, 2, 3, 10\}$ leads to degraded performance of ResNet-18 trained on ImageNet-1k for 150 epochs, in support of conclusion 2.

\bibliography{main}

@String(CVPR= {IEEE Conf. Comput. Vis. Pattern Recog.})

@String(ICCV= {Int. Conf. Comput. Vis.})

@String(ECCV= {Eur. Conf. Comput. Vis.})

@String(NIPS= {Adv. Neural Inform. Process. Syst.})

@String(ICLR = {Int. Conf. Learn. Represent.})

@String(CVPR  = {CVPR})

@String(ICCV  = {ICCV})

@String(ECCV  = {ECCV})

@String(NIPS  = {NeurIPS})

@String(ICLR  = {ICLR})

@INPROCEEDINGS {7298594,
author = { Szegedy, Christian and Wei Liu and Yangqing Jia and Sermanet, Pierre and Reed, Scott and Anguelov, Dragomir and Erhan, Dumitru and Vanhoucke, Vincent and Rabinovich, Andrew },
booktitle = { 2015 IEEE Conference on Computer Vision and Pattern Recognition (CVPR) },
title = { Going deeper with convolutions },
year = {2015},
volume = {},
ISSN = {1063-6919},
pages = {1-9},
doi = {10.1109/CVPR.2015.7298594},
url = {https://doi.ieeecomputersociety.org/10.1109/CVPR.2015.7298594},
publisher = {IEEE Computer Society},
address = {Los Alamitos, CA, USA},
month =Jun}

@inproceedings{
wightman2021resnet,
title={ResNet strikes back: An improved training procedure in timm},
author={Ross Wightman and Hugo Touvron and Herve Jegou},
booktitle={NeurIPS 2021 Workshop on ImageNet: Past, Present, and Future},
year={2021},
url={https://openreview.net/forum?id=NG6MJnVl6M5}
}

@INPROCEEDINGS{5206848,
  author={Deng, Jia and Dong, Wei and Socher, Richard and Li, Li-Jia and Kai Li and Li Fei-Fei},
  booktitle={2009 IEEE Conference on Computer Vision and Pattern Recognition}, 
  title={ImageNet: A large-scale hierarchical image database}, 
  year={2009},
  volume={},
  number={},
  pages={248-255},
  doi={10.1109/CVPR.2009.5206848},
  ISSN={1063-6919},
  month={June},}

@inproceedings{10.5555/3524938.3525087,
author = {Chen, Ting and Kornblith, Simon and Norouzi, Mohammad and Hinton, Geoffrey},
title = {A simple framework for contrastive learning of visual representations},
year = {2020},
publisher = {JMLR.org},
booktitle = {Proceedings of the 37th International Conference on Machine Learning},
articleno = {149},
numpages = {11},
series = {ICML'20}
}

@article{chen2020big,
  title={Big self-supervised models are strong semi-supervised learners},
  author={Chen, Ting and Kornblith, Simon and Swersky, Kevin and Norouzi, Mohammad and Hinton, Geoffrey E},
  journal={Advances in neural information processing systems},
  volume={33},
  pages={22243--22255},
  year={2020}
}

@article{Jordan202494OC,
  title={94\% on CIFAR-10 in 3.29 Seconds on a Single GPU},
  author={Keller Jordan},
  journal={ArXiv},
  year={2024},
  volume={abs/2404.00498},
  url={https://api.semanticscholar.org/CorpusID:268819769}
}

@article{
steiner2022how,
title={How to train your ViT? Data, Augmentation, and Regularization in Vision Transformers},
author={Andreas Peter Steiner and Alexander Kolesnikov and Xiaohua Zhai and Ross Wightman and Jakob Uszkoreit and Lucas Beyer},
journal={Transactions on Machine Learning Research},
issn={2835-8856},
year={2022},
url={https://openreview.net/forum?id=4nPswr1KcP},
note={}
}

@inproceedings{10.5555/3666122.3666890,
author = {Kirichenko, Polina and Ibrahim, Mark and Balestriero, Randall and Bouchacourt, Diane and Vedantam, Ramakrishna and Firooz, Hamed and Wilson, Andrew Gordon},
title = {Understanding the detrimental class-level effects of data augmentation},
year = {2023},
publisher = {Curran Associates Inc.},
address = {Red Hook, NY, USA},
booktitle = {Proceedings of the 37th International Conference on Neural Information Processing Systems},
articleno = {768},
numpages = {29},
location = {New Orleans, LA, USA},
series = {NIPS '23}
}

@article{beyer2022better,
  title={Better plain vit baselines for imagenet-1k},
  author={Beyer, Lucas and Zhai, Xiaohua and Kolesnikov, Alexander},
  journal={arXiv preprint arXiv:2205.01580},
  year={2022}
}

@misc{big_vision,
  author = {Beyer, Lucas and Zhai, Xiaohua and Kolesnikov, Alexander},
  title = {Big Vision},
  year = {2022},
  publisher = {GitHub},
  journal = {GitHub repository},
  howpublished = {\url{https://github.com/google-research/big_vision}}
}

@software{TorchVision_maintainers_and_contributors_TorchVision_PyTorch_s_Computer_2016,
author = {TorchVision maintainers and contributors},
license = {BSD-3-Clause},
month = nov,
title = {{TorchVision: PyTorch's Computer Vision library}},
url = {https://github.com/pytorch/vision},
year = {2016}
}

@inproceedings{zhai2022scaling,
  title={Scaling vision transformers},
  author={Zhai, Xiaohua and Kolesnikov, Alexander and Houlsby, Neil and Beyer, Lucas},
  booktitle={Proceedings of the IEEE/CVF conference on computer vision and pattern recognition},
  pages={12104--12113},
  year={2022}
}

@inproceedings{beyer2023flexivit,
  title={Flexivit: One model for all patch sizes},
  author={Beyer, Lucas and Izmailov, Pavel and Kolesnikov, Alexander and Caron, Mathilde and Kornblith, Simon and Zhai, Xiaohua and Minderer, Matthias and Tschannen, Michael and Alabdulmohsin, Ibrahim and Pavetic, Filip},
  booktitle={Proceedings of the IEEE/CVF Conference on Computer Vision and Pattern Recognition},
  pages={14496--14506},
  year={2023}
}

@article{dehghani2024patch,
  title={Patch n’pack: Navit, a vision transformer for any aspect ratio and resolution},
  author={Dehghani, Mostafa and Mustafa, Basil and Djolonga, Josip and Heek, Jonathan and Minderer, Matthias and Caron, Mathilde and Steiner, Andreas and Puigcerver, Joan and Geirhos, Robert and Alabdulmohsin, Ibrahim M and others},
  journal={Advances in Neural Information Processing Systems},
  volume={36},
  year={2024}
}

@article{alabdulmohsin2024getting,
  title={Getting ViT in shape: Scaling laws for compute-optimal model design},
  author={Alabdulmohsin, Ibrahim M and Zhai, Xiaohua and Kolesnikov, Alexander and Beyer, Lucas},
  journal={Advances in Neural Information Processing Systems},
  volume={36},
  year={2024}
}

@Misc{Dahl2023AlgoPerf,
  title         = {Benchmarking Neural Network Training Algorithms},
  author        = {Dahl, George E. and Schneider, Frank and Nado, Zachary and Agarwal, Naman and Sastry, Chandramouli Shama and Hennig, Philipp and Medapati, Sourabh and Eschenhagen, Runa and Kasimbeg, Priya and Suo, Daniel and Bae, Juhan and Gilmer, Justin and Peirson, Abel L. and Khan, Bilal and Anil, Rohan and Rabbat, Mike and Krishnan, Shankar and Snider, Daniel and Amid, Ehsan and Chen, Kongtao and Maddison, Chris J. and Vasudev, Rakshith and Badura, Michal and Garg, Ankush and Mattson, Peter},
  year          = {2023},
  archiveprefix = {arXiv},
  eprint        = {2306.07179},
}

@InProceedings{10.1007/978-3-031-73661-2_5,
author="Qin, Danfeng
and Leichner, Chas
and Delakis, Manolis
and Fornoni, Marco
and Luo, Shixin
and Yang, Fan
and Wang, Weijun
and Banbury, Colby
and Ye, Chengxi
and Akin, Berkin
and Aggarwal, Vaibhav
and Zhu, Tenghui
and Moro, Daniele
and Howard, Andrew",
editor="Leonardis, Ale{\v{s}}
and Ricci, Elisa
and Roth, Stefan
and Russakovsky, Olga
and Sattler, Torsten
and Varol, G{\"u}l",
title="MobileNetV4: Universal Models for the Mobile Ecosystem",
booktitle="Computer Vision -- ECCV 2024",
year="2025",
publisher="Springer Nature Switzerland",
address="Cham",
pages="78--96",
isbn="978-3-031-73661-2"
}

@inproceedings{NEURIPS2021_a2fe8c05,
 author = {Bouchacourt, Diane and Ibrahim, Mark and Morcos, Ari},
 booktitle = {Advances in Neural Information Processing Systems},
 editor = {M. Ranzato and A. Beygelzimer and Y. Dauphin and P.S. Liang and J. Wortman Vaughan},
 pages = {19566--19579},
 publisher = {Curran Associates, Inc.},
 title = {Grounding inductive biases in natural images: invariance stems from variations in data},
 url = {https://proceedings.neurips.cc/paper_files/paper/2021/file/a2fe8c05877ec786290dd1450c3385cd-Paper.pdf},
 volume = {34},
 year = {2021}
}

@inproceedings{
dosovitskiy2021an,
title={An Image is Worth 16x16 Words: Transformers for Image Recognition at Scale},
author={Alexey Dosovitskiy and Lucas Beyer and Alexander Kolesnikov and Dirk Weissenborn and Xiaohua Zhai and Thomas Unterthiner and Mostafa Dehghani and Matthias Minderer and Georg Heigold and Sylvain Gelly and Jakob Uszkoreit and Neil Houlsby},
booktitle={International Conference on Learning Representations},
year={2021},
url={https://openreview.net/forum?id=YicbFdNTTy}
}

@inproceedings{
raghu2021do,
title={Do Vision Transformers See Like Convolutional Neural Networks?},
author={Maithra Raghu and Thomas Unterthiner and Simon Kornblith and Chiyuan Zhang and Alexey Dosovitskiy},
booktitle={Advances in Neural Information Processing Systems},
editor={A. Beygelzimer and Y. Dauphin and P. Liang and J. Wortman Vaughan},
year={2021},
url={https://openreview.net/forum?id=R-616EWWKF5}
}

@INPROCEEDINGS{9711302,
  author={Chen, Xinlei and Xie, Saining and He, Kaiming},
  booktitle={2021 IEEE/CVF International Conference on Computer Vision (ICCV)}, 
  title={An Empirical Study of Training Self-Supervised Vision Transformers}, 
  year={2021},
  volume={},
  number={},
  pages={9620-9629},
  doi={10.1109/ICCV48922.2021.00950}}

@inproceedings{NEURIPS2020_d85b63ef,
 author = {Cubuk, Ekin Dogus and Zoph, Barret and Shlens, Jon and Le, Quoc},
 booktitle = {Advances in Neural Information Processing Systems},
 editor = {H. Larochelle and M. Ranzato and R. Hadsell and M.F. Balcan and H. Lin},
 pages = {18613--18624},
 publisher = {Curran Associates, Inc.},
 title = {RandAugment: Practical Automated Data Augmentation with a Reduced Search Space},
 url = {https://proceedings.neurips.cc/paper_files/paper/2020/file/d85b63ef0ccb114d0a3bb7b7d808028f-Paper.pdf},
 volume = {33},
 year = {2020}
}

@inproceedings{
zhang2018mixup,
title={mixup: Beyond Empirical Risk Minimization},
author={Hongyi Zhang and Moustapha Cisse and Yann N. Dauphin and David Lopez-Paz},
booktitle={International Conference on Learning Representations},
year={2018},
url={https://openreview.net/forum?id=r1Ddp1-Rb},
}

@misc{
hernandez-garcia2020data,
title={Data augmentation instead of explicit regularization},
author={Alex Hernandez-Garcia and Peter K{\"o}nig},
year={2020},
url={https://openreview.net/forum?id=H1eqOnNYDH}
}

@inproceedings{NIPS1991_65658fde,
 author = {Simard, Patrice and Victorri, Bernard and LeCun, Yann and Denker, John},
 booktitle = {Advances in Neural Information Processing Systems},
 editor = {J. Moody and S. Hanson and R.P. Lippmann},
 pages = {},
 publisher = {Morgan-Kaufmann},
 title = {Tangent Prop - A formalism for specifying selected invariances in an adaptive network},
 url = {https://proceedings.neurips.cc/paper_files/paper/1991/file/65658fde58ab3c2b6e5132a39fae7cb9-Paper.pdf},
 volume = {4},
 year = {1991}
}

@INPROCEEDINGS {9711158,
author = { Muller, Samuel G. and Hutter, Frank },
booktitle = { 2021 IEEE/CVF International Conference on Computer Vision (ICCV) },
title = {{ TrivialAugment: Tuning-free Yet State-of-the-Art Data Augmentation }},
year = {2021},
volume = {},
ISSN = {},
pages = {754-762},
doi = {10.1109/ICCV48922.2021.00081},
url = {https://doi.ieeecomputersociety.org/10.1109/ICCV48922.2021.00081},
publisher = {IEEE Computer Society},
address = {Los Alamitos, CA, USA},
month =Oct}

@INPROCEEDINGS{9008296,
  author={Yun, Sangdoo and Han, Dongyoon and Chun, Sanghyuk and Oh, Seong Joon and Yoo, Youngjoon and Choe, Junsuk},
  booktitle={2019 IEEE/CVF International Conference on Computer Vision (ICCV)}, 
  title={CutMix: Regularization Strategy to Train Strong Classifiers With Localizable Features}, 
  year={2019},
  volume={},
  number={},
  pages={6022-6031},
  doi={10.1109/ICCV.2019.00612}}

@inproceedings{
hendrycks*2020augmix,
title={AugMix: A Simple Method to Improve Robustness and Uncertainty under Data Shift},
author={Dan Hendrycks* and Norman Mu* and Ekin Dogus Cubuk and Barret Zoph and Justin Gilmer and Balaji Lakshminarayanan},
booktitle={International Conference on Learning Representations},
year={2020},
url={https://openreview.net/forum?id=S1gmrxHFvB}
}

@article{russakovsky2015imagenet,
  title={Imagenet large scale visual recognition challenge},
  author={Russakovsky, Olga and Deng, Jia and Su, Hao and Krause, Jonathan and Satheesh, Sanjeev and Ma, Sean and Huang, Zhiheng and Karpathy, Andrej and Khosla, Aditya and Bernstein, Michael and others},
  journal={International journal of computer vision},
  volume={115},
  pages={211--252},
  year={2015},
  publisher={Springer}
}

@inproceedings{
ridnik2021imagenetk,
title={ImageNet-21K Pretraining for the Masses},
author={Tal Ridnik and Emanuel Ben-Baruch and Asaf Noy and Lihi Zelnik-Manor},
booktitle={Thirty-fifth Conference on Neural Information Processing Systems Datasets and Benchmarks Track (Round 1)},
year={2021},
url={https://openreview.net/forum?id=Zkj_VcZ6ol}
}

@article{kisel2024flaws,
  title={Flaws of ImageNet, Computer Vision's Favourite Dataset},
  author={Kisel, Nikita and Volkov, Illia and Hanzelkova, Katerina and Janouskova, Klara and Matas, Jiri},
  journal={arXiv preprint arXiv:2412.00076},
  year={2024}
}

@inproceedings{
vasudevan2022when,
title={When does dough become a bagel? Analyzing the remaining mistakes on ImageNet},
author={Vijay Vasudevan and Benjamin Caine and Raphael Gontijo-Lopes and Sara Fridovich-Keil and Rebecca Roelofs},
booktitle={Advances in Neural Information Processing Systems},
editor={Alice H. Oh and Alekh Agarwal and Danielle Belgrave and Kyunghyun Cho},
year={2022},
url={https://openreview.net/forum?id=mowt1WNhTC7}
}

@INPROCEEDINGS{9577610,
  author={Yun, Sangdoo and Oh, Seong Joon and Heo, Byeongho and Han, Dongyoon and Choe, Junsuk and Chun, Sanghyuk},
  booktitle={2021 IEEE/CVF Conference on Computer Vision and Pattern Recognition (CVPR)}, 
  title={Re-labeling ImageNet: from Single to Multi-Labels, from Global to Localized Labels}, 
  year={2021},
  volume={},
  number={},
  pages={2340-2350},
  doi={10.1109/CVPR46437.2021.00237},
  ISSN={2575-7075},
  month={June},}

@article{beyer2020we,
  title={Are we done with imagenet?},
  author={Beyer, Lucas and H{\'e}naff, Olivier J and Kolesnikov, Alexander and Zhai, Xiaohua and Oord, A{\"a}ron van den},
  journal={arXiv preprint arXiv:2006.07159},
  year={2020}
}

@InProceedings{dehghani2021scenic,
    author    = {Dehghani, Mostafa and Gritsenko, Alexey and Arnab, Anurag and Minderer, Matthias and Tay, Yi},
    title     = {Scenic: A JAX Library for Computer Vision Research and Beyond},
    booktitle = {Proceedings of the IEEE/CVF Conference on Computer Vision and Pattern Recognition (CVPR)},
    year      = {2022},
    pages     = {21393-21398}
}

@inproceedings{Ansel_PyTorch_2_Faster_2024,
author = {Ansel, Jason and Yang, Edward and He, Horace and Gimelshein, Natalia and Jain, Animesh and Voznesensky, Michael and Bao, Bin and Bell, Peter and Berard, David and Burovski, Evgeni and Chauhan, Geeta and Chourdia, Anjali and Constable, Will and Desmaison, Alban and DeVito, Zachary and Ellison, Elias and Feng, Will and Gong, Jiong and Gschwind, Michael and Hirsh, Brian and Huang, Sherlock and Kalambarkar, Kshiteej and Kirsch, Laurent and Lazos, Michael and Lezcano, Mario and Liang, Yanbo and Liang, Jason and Lu, Yinghai and Luk, CK and Maher, Bert and Pan, Yunjie and Puhrsch, Christian and Reso, Matthias and Saroufim, Mark and Siraichi, Marcos Yukio and Suk, Helen and Suo, Michael and Tillet, Phil and Wang, Eikan and Wang, Xiaodong and Wen, William and Zhang, Shunting and Zhao, Xu and Zhou, Keren and Zou, Richard and Mathews, Ajit and Chanan, Gregory and Wu, Peng and Chintala, Soumith},
booktitle = {29th ACM International Conference on Architectural Support for Programming Languages and Operating Systems, Volume 2 (ASPLOS '24)},
doi = {10.1145/3620665.3640366},
month = apr,
publisher = {ACM},
title = {{PyTorch 2: Faster Machine Learning Through Dynamic Python Bytecode Transformation and Graph Compilation}},
url = {https://pytorch.org/assets/pytorch2-2.pdf},
year = {2024}
}

@inproceedings{springenberg2015striving,
  title={Striving for Simplicity: The All Convolutional Net},
  author={Springenberg, J and Dosovitskiy, Alexey and Brox, Thomas and Riedmiller, M},
  booktitle={ICLR (workshop track)},
  year={2015}
}

@inproceedings{zhou2016learning,
  title={Learning deep features for discriminative localization},
  author={Zhou, Bolei and Khosla, Aditya and Lapedriza, Agata and Oliva, Aude and Torralba, Antonio},
  booktitle={Proceedings of the IEEE conference on computer vision and pattern recognition},
  pages={2921--2929},
  year={2016}
}

@inproceedings{selvaraju2017grad,
  title={Grad-cam: Visual explanations from deep networks via gradient-based localization},
  author={Selvaraju, Ramprasaath R and Cogswell, Michael and Das, Abhishek and Vedantam, Ramakrishna and Parikh, Devi and Batra, Dhruv},
  booktitle={Proceedings of the IEEE international conference on computer vision},
  pages={618--626},
  year={2017}
}

@inproceedings{binder2016layer,
  title={Layer-wise relevance propagation for neural networks with local renormalization layers},
  author={Binder, Alexander and Montavon, Gr{\'e}goire and Lapuschkin, Sebastian and M{\"u}ller, Klaus-Robert and Samek, Wojciech},
  booktitle={Artificial Neural Networks and Machine Learning--ICANN 2016: 25th International Conference on Artificial Neural Networks, Barcelona, Spain, September 6-9, 2016, Proceedings, Part II 25},
  pages={63--71},
  year={2016},
  organization={Springer}
}

@inproceedings{chefer2021transformer,
  title={Transformer interpretability beyond attention visualization},
  author={Chefer, Hila and Gur, Shir and Wolf, Lior},
  booktitle={Proceedings of the IEEE/CVF conference on computer vision and pattern recognition},
  pages={782--791},
  year={2021}
}

@inproceedings{hagele2024scaling,
  title={Scaling Laws and Compute-Optimal Training Beyond Fixed Training Durations},
  author={Alexander H{\"a}gele and Elie Bakouch and Atli Kosson and Loubna Ben allal and Leandro Von Werra and Martin Jaggi},
  booktitle={The Thirty-eighth Annual Conference on Neural Information Processing Systems},
  year={2024},
  url={https://openreview.net/forum?id=Y13gSfTjGr}
}

@inproceedings{
defazio2024the,
title={The Road Less Scheduled},
author={Aaron Defazio and Xingyu Alice Yang and Ahmed Khaled and Konstantin Mishchenko and Harsh Mehta and Ashok Cutkosky},
booktitle={The Thirty-eighth Annual Conference on Neural Information Processing Systems},
year={2024},
url={https://openreview.net/forum?id=0XeNkkENuI}
}

@ARTICLE{10530470,
  author={Wang, Yulin and Yue, Yang and Lu, Rui and Han, Yizeng and Song, Shiji and Huang, Gao},
  journal={IEEE Transactions on Pattern Analysis and Machine Intelligence}, 
  title={EfficientTrain++: Generalized Curriculum Learning for Efficient Visual Backbone Training}, 
  year={2024},
  volume={46},
  number={12},
  pages={8036-8055},
  doi={10.1109/TPAMI.2024.3401036}}

@inproceedings{
bayat2025the,
title={The Pitfalls of Memorization: When Memorization Hurts Generalization},
author={Reza Bayat and Mohammad Pezeshki and Elvis Dohmatob and David Lopez-Paz and Pascal Vincent},
booktitle={The Thirteenth International Conference on Learning Representations},
year={2025},
url={https://openreview.net/forum?id=vVhZh9ZpIM}
}

@inproceedings{
loshchilov2018decoupled,
title={Decoupled Weight Decay Regularization},
author={Ilya Loshchilov and Frank Hutter},
booktitle={International Conference on Learning Representations},
year={2019},
url={https://openreview.net/forum?id=Bkg6RiCqY7},
}

@inproceedings{
pethick2025training,
title={Training Deep Learning Models with Norm-Constrained {LMO}s},
author={Thomas Pethick and Wanyun Xie and Kimon Antonakopoulos and Zhenyu Zhu and Antonio Silveti-Falls and Volkan Cevher},
booktitle={Forty-second International Conference on Machine Learning},
year={2025},
url={https://openreview.net/forum?id=2Oqm2IzTy9}
}

@InProceedings{pmlr-v139-touvron21a,
  title = 	 {Training data-efficient image transformers \& distillation through attention},
  author =       {Touvron, Hugo and Cord, Matthieu and Douze, Matthijs and Massa, Francisco and Sablayrolles, Alexandre and Jegou, Herve},
  booktitle = 	 {Proceedings of the 38th International Conference on Machine Learning},
  pages = 	 {10347--10357},
  year = 	 {2021},
  editor = 	 {Meila, Marina and Zhang, Tong},
  volume = 	 {139},
  series = 	 {Proceedings of Machine Learning Research},
  month = 	 {18--24 Jul},
  publisher =    {PMLR},
  url = 	 {https://proceedings.mlr.press/v139/touvron21a.html}
}

@misc{vit-pytorch,
author = {Wang, Phil},
title = {vit-pytorch},
url = {https://github.com/lucidrains/vit-pytorch}
}
\bibliographystyle{tmlr}

\appendix

\section{TensorFlow vs. PyTorch Inception crop training loss}
\begin{figure}[H]
  \centering
  \includegraphics[width=0.8\linewidth]{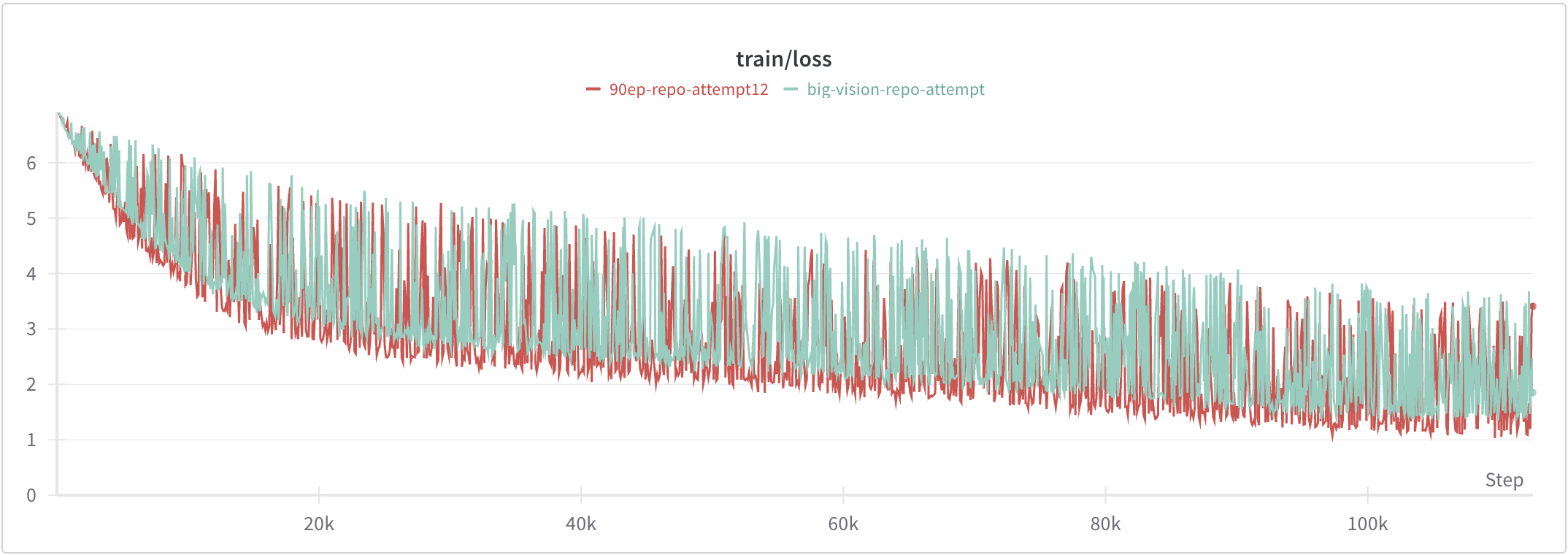}
  \caption{Training loss of the linear head variant of ViT-S/16 from \cite{beyer2022better} for 90 epochs using the Big Vision \citep{big_vision} first-party code (cyan) vs. our reimplementation in TorchVision \citep{TorchVision_maintainers_and_contributors_TorchVision_PyTorch_s_Computer_2016} (red).}
  \label{fig:training_loss_discrepancy}
\end{figure}

\Cref{fig:training_loss_discrepancy} shows the training loss discrepancy when we train the linear head variant of ViT-S/16 from \cite{beyer2022better} for 90 epochs. Since the TensorFlow Inception crop used by Big Vision oversamples smaller crops, it consistently reports higher training loss.

\label{sec:training_loss}
\section{Crop size distribution of the PyTorch Inception crop}
\label{sec:torch_implementation}
\begin{figure}[H]
    \centering
    \includegraphics[width=0.9\columnwidth]{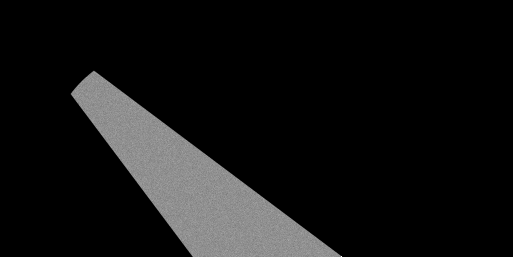}
    \caption{Probability heatmap of the crop size $(h, w)$ given by the PyTorch Inception crop (\texttt{RandomResizedCrop}, RRC) for landscape image ($h_o=256, w_o=512$) with the same parameters as in \cref{fig:tf_heat_map}: $r_\mathrm{min} = \frac{3}{4}, r_\mathrm{max} = \frac{4}{3}, a_\mathrm{min} = 0.05, a_\mathrm{max} = 1, N=10^7$. The counterpart for portrait image ($h_o=512, w_o=256$ is indistinguishable after image transposition.}
    \label{fig:torch_heat_map}
\end{figure}

Here is the probability heatmap of the crop size $(h, w)$ given by the PyTorch Inception crop for landscape and portrait images (\cref{fig:torch_heat_map}). About $1.4 \times 10^4$ out of $N=10^7$ calls of RRC fail to sample a constraint-respecting crop within the hard-coded \texttt{max\_attempts=10} and fall back to central crop, but otherwise the distribution is uniform and equivariant to image transposition.

\section{Beta crop implementation}
\label{sec:beta_implementation}

The source code of our implementation of Beta crop is available \href{https://github.com/EIFY/mup-vit/blob/beta-crop/transforms.py#L46}{here} which follows \cref{alg:beta_crop}. The part that differs from \cref{alg:torch_crop} is highlighted in \textcolor{red}{red}. Clipping $A_\mathrm{max}$ reduces sampling rejections and makes sure that we use the full support $x \in [0, 1]$ of the beta distribution.

\begin{algorithm}[H]
\begin{algorithmic}[1]
    \State {\bfseries Input:} img, \textcolor{red}{$\alpha, \beta$}, $r_\mathrm{min}, r_\mathrm{max}, a_\mathrm{min}, a_\mathrm{max}$
    \State $\_, h_o, w_o$ = \texttt{get\_dimensions(}img\texttt{)}
    \State \textcolor{red}{$A_\mathrm{min} = a_\mathrm{min}h_ow_o$}
    \State \textcolor{red}{$A_\mathrm{max} = a_\mathrm{max}h_ow_o$}
    \If{\textcolor{red}{$r_o > r_\mathrm{max}$}}
        \State \textcolor{red}{$A_\mathrm{max} = \min(A_\mathrm{max}, r_\mathrm{max}h_o^2)$}
    \ElsIf{\textcolor{red}{$r_o < r_\mathrm{min}$}}
        \State \textcolor{red}{$A_\mathrm{max} = \min(A_\mathrm{max}, \frac{w_o^2}{r_\mathrm{min}})$}
    \EndIf
    \For{$t=1$ {\bfseries to} \textcolor{red}{\text{max\_attempts}}}
        \State Sample $r_\mathrm{log} \sim \mathcal{U}(\log(r_\mathrm{min}), \log(r_\mathrm{max}))$
        \State $r = e^{r_\mathrm{log}}$ 
        \State \textcolor{red}{Sample $x \sim Beta(\alpha, \beta)$}
        \State \textcolor{red}{$a = A_\mathrm{min} + x (A_\mathrm{max} - A_\mathrm{min})$}
        \State $w = \sqrt{ar}$
        \State $h = \sqrt{\frac{a}{r}}$    
        \If{$0 < h \le h_o$ and $0 < w \le w_o$}
            \State Sample $y \sim \mathcal{U}(0, h_o - h)$
            \State Sample $x \sim \mathcal{U}(0, w_o - w)$
            \State Return $(y, x, h, w)$
        \EndIf
    \EndFor
    \If{$r_o < r_\mathrm{min}$}
        \State $w = w_o$
        \State $h = \frac{w}{r_\mathrm{min}}$
    \ElsIf{$r_o > r_\mathrm{max}$}
        \State $h = h_o$
        \State $w = hr_\mathrm{max}$
    \Else
        \State $w = w_o$
        \State $h = h_o$
    \EndIf
    \State Return $(\frac{h_o - h}{2}, \frac{w_o - w}{2}, h, w)$ \Comment{Fallback to central crop}
\end{algorithmic}
\caption{\texttt{BetaCrop}}
\label{alg:beta_crop}
\end{algorithm}
\section{RandAugment implementation}
\label{sec:randaugment_implementation}
In contrast to the original RandAugment paper \citep{NEURIPS2020_d85b63ef} which describes RandAugment as having the following 14 transforms:
\begin{multicols}{3}
\begin{itemize}
    \par 
    \begingroup
    \nolinenumbers
    \item Identity
    \item AutoContrast
    \item Equalize
    \item Rotate
    \item Solarize
    \item Color
    \item Posterize
    \item Contrast
    \item Brightness
    \item Sharpness
    \par 
    \endgroup
    \item ShearX
    \item ShearY
    \item TranslateX
    \item TranslateY
\end{itemize}
\end{multicols}
The \texttt{efficientnet} \href{https://github.com/tensorflow/tpu/tree/master/models/official/efficientnet}{repository} linked from the paper implements RandAugment with a lineup of 16 transforms:
\begin{multicols}{3}
\begin{itemize}
    \item AutoContrast
    \item Equalize
    \item Rotate
    \item Solarize
    \item Color
    \item Posterize
    \par 
    \begingroup
    \nolinenumbers
    \item Contrast
    \item Brightness
    \item Sharpness
    \item ShearX
    \item ShearY
    \item TranslateX
    \item TranslateY
    \item Invert
    \item Cutout
    \item SolarizeAdd
    \par 
    \endgroup
\end{itemize}
\end{multicols}

\begin{figure}
    \centering
    \begin{subfigure}{0.33\textwidth}
    \centering
    \captionsetup{justification=centering}
      \includegraphics[width=\textwidth]{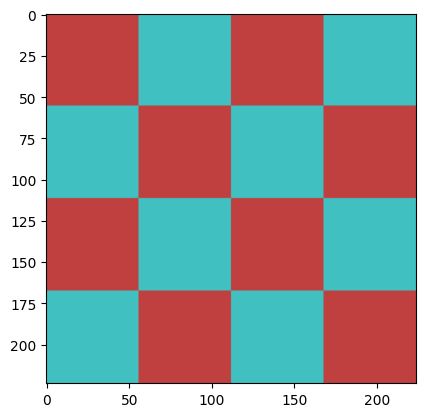}
      \caption*{Before transform: (192, 64, 64) and\\ (64, 192, 192)}
    \end{subfigure}%
    \begin{subfigure}{0.33\textwidth}
    \centering
    \captionsetup{justification=centering}
      \includegraphics[width=\textwidth]{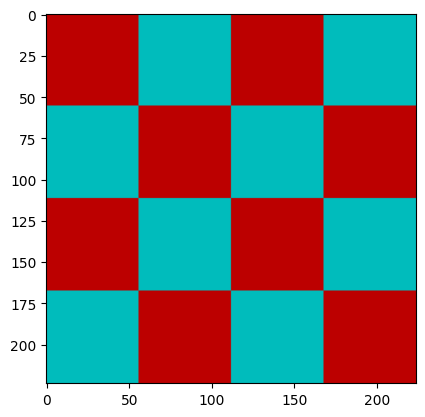}
      \caption*{Broken result: (188, 0, 0) and\\ (0, 188, 188)}
    \end{subfigure}%
    \begin{subfigure}{0.33\textwidth}
    \centering
    \captionsetup{justification=centering}
      \includegraphics[width=\textwidth]{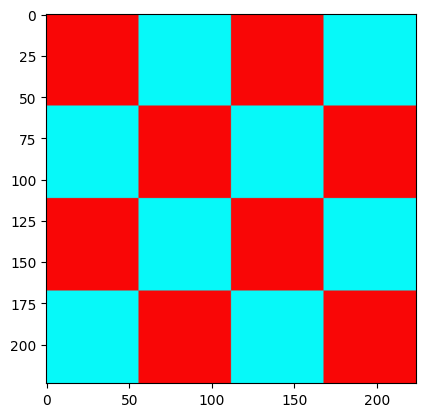}
      \caption*{Correct result: (249, 6, 6) and\\ (6, 249, 249)}
    \end{subfigure}%
    \caption{Color grids, before and after calling \texttt{Contrast($\cdot$, 1.9)} and their RGB values.}
    \label{fig:broken_contrast}
\end{figure}

In addition, the \texttt{Contrast} transform is broken. In short, when trying to compute the average RGB values, it accidentally computes $\frac{hw}{256}$ ,where $hw$ is the number of pixels, instead of the weighted average of the binned values. As a result, it tends to either brighten or darken the image instead of enhancing or reducing the contrast (\cref{fig:broken_contrast}). We reimplement the 16-transform lineup following the source code of the repository but fix the \texttt{Contrast} transform in PyTorch (\href{https://github.com/EIFY/mup-vit/blob/beta-crop/main.py#L337}{source code}).


\section{Reimplementation of the TensorFlow Inception crop in PyTorch}
\label{sec:torch_reimplementation}
\begin{figure}[H]
  \centering
  \includegraphics[width=0.4\linewidth]{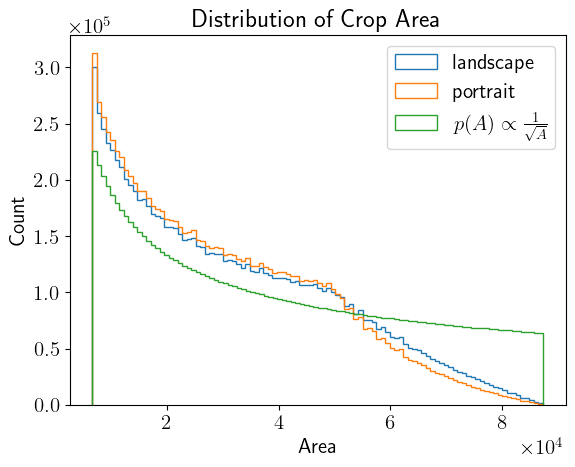}
  \caption{Crop area distribution of the reimplementation of TensorFlow Inception crop in PyTorch with default parameters for $256 \times 512$ and $512 \times 256$ images, $N=10^7$. The results are indistinguishable from \cref{fig:tf_area_distribution_approx}.}
  \label{fig:torch_area_distribution_approx}
\end{figure}

For convenience and consistency, we reimplement the TensorFlow Inception crop in PyTorch (\href{https://github.com/EIFY/mup-vit/blob/beta-crop/transforms.py#L131}{source code}) and use it in our experiments instead of switching to TensorFlow or Jax. We have verified that both the crop area distribution (\cref{fig:torch_area_distribution_approx}) and the probability heat maps are indistinguishable from that of the original TensorFlow Inception crop.
\section{Detailed Inception crop results}
\label{sec:inception_full}

\subsection{Full tables with error bars}
\label{sec:inception_error_bars}
\Cref{table:tf_original_full,table:torch_original_full,table:tf_real_full,table:torch_real_full} show the full results for training ViT-S/16 models on ImageNet-1k with implementations of Inception crop, varying training budget and $a_\mathrm{min}$. We train $N=3$ for each setting and values are given as (mean) $\pm$ (sample standard deviation), both rounded to 2 decimal places.

\begin{table*}
\small
\centering
\begin{tabular}{|c||c|c|c|c|c|c|c|c|c|c|c|c|}
\hline
$a_\mathrm{min}$ & 0.025 & 0.05 & 0.1 & 0.15 & 0.2 & 0.25 \\
\hline
30ep & 64.59$\pm 0.11$ & 65.55$\pm 0.07$ & 67.03$\pm 0.36$ & 68.06$\pm 0.19$ & 68.78$\pm 0.05$ & \textbf{69.15}$\pm 0.06$ \\
\hline
60ep & 72.77$\pm 0.21$ & 73.43$\pm 0.13$ & 74.29$\pm 0.15$ & 74.77$\pm 0.05$ & 74.99$\pm 0.08$ & \textbf{75.25}$\pm 0.10$ \\
\hline
90ep & 75.76$\pm 0.04$ & 76.25$\pm 0.05$ & 76.77$\pm 0.09$ & 77.03$\pm 0.08$ & \textbf{77.21}$\pm 0.13$ & 77.16$\pm 0.16$ \\
\hline
150ep & 77.90$\pm 0.07$ & 78.29$\pm 0.06$ & \textbf{78.60}$\pm 0.06$ & 78.59$\pm 0.10$ & 78.52$\pm 0.06$ & 78.39$\pm 0.13$ \\
\hline
300ep & 79.43$\pm 0.07$ & \textbf{79.75}$\pm 0.07$ & 79.64$\pm 0.14$ & 79.49$\pm 0.12$ & 79.20$\pm 0.10$ & 79.01$\pm 0.14$ \\
\hline
\end{tabular}
\caption{Top-1 val. accuracy (original label), TensorFlow Inception crop. As the training budget increases from 30 epochs to 300 epochs, the optimal crop scale lower bound $a_\mathrm{min}$ decreases (stronger augmentation).}
\label{table:tf_original_full}
\end{table*}

\begin{table*}
\small
\centering
\begin{tabular}{|c||c|c|c|c|c|c|c|c|c|c|c|c|}
\hline
$a_\mathrm{min}$ & 0.025 & 0.05 & 0.1 & 0.15 & 0.2 & 0.25 \\
\hline
30ep & 67.08$\pm 0.41$ & 67.35$\pm 0.33$ & 68.26$\pm 0.20$ & 68.74$\pm 0.20$ & 69.26$\pm 0.13$ & \textbf{69.67}$\pm 0.17$ \\
\hline
60ep & 74.35$\pm 0.08$ & 74.77$\pm 0.08$ & 74.86$\pm 0.13$ & 75.08$\pm 0.22$ & 75.20$\pm 0.07$ & \textbf{75.55}$\pm 0.11$ \\
\hline
90ep & 76.81$\pm 0.16$ & 76.92$\pm 0.13$ & 77.21$\pm 0.19$ & 77.21$\pm 0.17$ & \textbf{77.35}$\pm 0.14$ & 77.16$\pm 0.09$ \\
\hline
150ep & 78.58$\pm 0.02$ & 78.64$\pm 0.18$ & \textbf{78.75}$\pm 0.04$ & 78.70$\pm 0.10$ & 78.51$\pm 0.17$ & 78.43$\pm 0.11$ \\
\hline
300ep & \textbf{79.80}$\pm 0.06$ & 79.73$\pm 0.12$ & 79.62$\pm 0.05$ & 79.18$\pm 0.03$ & 79.13$\pm 0.04$ & 78.86$\pm 0.01$ \\
\hline
\end{tabular}
\caption{Top-1 val. accuracy (original label), PyTorch Inception crop. As the training budget increases from 30 epochs to 300 epochs, the optimal crop scale lower bound $a_\mathrm{min}$ decreases (stronger augmentation).}
\label{table:torch_original_full}
\end{table*}

\begin{table*}
\small
\centering
\begin{tabular}{|c||c|c|c|c|c|c|c|c|c|c|c|c|}
\hline
$a_\mathrm{min}$ & 0.025 & 0.05 & 0.1 & 0.15 & 0.2 & 0.25 \\
\hline
30ep & 72.23$\pm 0.13$ & 73.22$\pm 0.18$ & 74.77$\pm 0.39$ & 75.73$\pm 0.28$ & 76.45$\pm 0.02$ & \textbf{76.84}$\pm 0.11$ \\
\hline
60ep & 80.18$\pm 0.23$ & 80.75$\pm 0.02$ & 81.30$\pm 0.15$ & 81.68$\pm 0.08$ & 81.88$\pm 0.08$ & \textbf{82.13}$\pm 0.03$ \\
\hline
90ep & 82.65$\pm 0.05$ & 83.10$\pm 0.03$ & 83.31$\pm 0.08$ & 83.57$\pm 0.06$ & \textbf{83.64}$\pm 0.07$ & 83.56$\pm 0.16$ \\
\hline
150ep & 84.21$\pm 0.06$ & 84.51$\pm 0.10$ & \textbf{84.67}$\pm 0.03$ & 84.51$\pm 0.08$ & 84.42$\pm 0.07$ & 84.36$\pm 0.10$ \\
\hline
300ep & 85.30$\pm 0.09$ & \textbf{85.39}$\pm 0.09$ & 85.25$\pm 0.11$ & 85.01$\pm 0.07$ & 84.71$\pm 0.09$ & 84.45$\pm 0.19$ \\
\hline
\end{tabular}
\caption{Top-1 val. accuracy (ReaL), TensorFlow Inception crop. As the training budget increases from 30 epochs to 300 epochs, the optimal crop scale lower bound $a_\mathrm{min}$ decreases (stronger augmentation).}
\label{table:tf_real_full}
\end{table*}

\begin{table*}
\small
\centering
\begin{tabular}{|c||c|c|c|c|c|c|c|c|c|c|c|c|}
\hline
$a_\mathrm{min}$ & 0.025 & 0.05 & 0.1 & 0.15 & 0.2 & 0.25 \\
\hline
30ep & 74.77$\pm 0.33$ & 75.11$\pm 0.32$ & 75.90$\pm 0.18$ & 76.39$\pm 0.14$ & 76.88$\pm 0.10$ & \textbf{77.21}$\pm 0.09$ \\
\hline
60ep & 81.44$\pm 0.11$ & 81.73$\pm 0.09$ & 81.86$\pm 0.12$ & 81.95$\pm 0.11$ & 82.09$\pm 0.09$ & \textbf{82.37}$\pm 0.05$ \\
\hline
90ep & 83.39$\pm 0.15$ & 83.39$\pm 0.09$ & \textbf{83.65}$\pm 0.14$ & 83.58$\pm 0.17$ & 83.59$\pm 0.07$ & 83.50$\pm 0.08$ \\
\hline
150ep & 84.57$\pm 0.14$ & \textbf{84.65}$\pm 0.17$ & 84.56$\pm 0.05$ & 84.47$\pm 0.08$ & 84.33$\pm 0.06$ & 84.28$\pm 0.07$ \\
\hline
300ep & \textbf{85.32}$\pm 0.15$ & 85.27$\pm 0.03$ & 85.06$\pm 0.01$ & 84.81$\pm 0.09$ & 84.59$\pm 0.06$ & 84.44$\pm 0.08$ \\
\hline
\end{tabular}
\caption{Top-1 val. accuracy (ReaL), PyTorch Inception crop. As the training budget increases from 30 epochs to 300 epochs, the optimal crop scale lower bound $a_\mathrm{min}$ decreases (stronger augmentation).}
\label{table:torch_real_full}
\end{table*}

\subsection{Statistical significance}
\label{sec:inception_statistical_significance}
\Cref{table:t_test_tf_vs_torch_original,table:t_test_tf_vs_torch_real} show the statistical significance of the differences in top-1 val. accuracies between the TensorFlow and PyTorch Inception crop with various training budgets while \cref{table:t_test_tf_trend_original,table:t_test_tf_trend_real} show the statistical significance of the differences in top-1 val. accuracies across different values of crop scale lower bound $a_{min}$ for the TensorFlow Inception crop and \cref{table:t_test_torch_trend_original,table:t_test_torch_trend_real} show the statistical significance of the differences in top-1 val. accuracies across different values of crop scale lower bound $a_{min}$ for the PyTorch Inception crop.

\begin{table}[hbt!]
\small
\centering
\begin{tabular}{|c||c|c|c|c|c|c|}
\hline
$a_{min}$ & 0.025 & 0.05 & 0.1 & 0.15 & 0.2 & 0.25 \\
\hline
30ep & \cellcolor{red!45}5.1e-4 & \cellcolor{red!45}8.0e-4 & \cellcolor{red!30}6.6e-3 & \cellcolor{red!15}0.013 & \cellcolor{red!30}3.3e-3 & \cellcolor{red!30}7.5e-3 \\
\hline
60ep & \cellcolor{red!45}2.6e-4 & \cellcolor{red!45}1.e-4 & \cellcolor{red!30}7.3e-3 & 0.074 & \cellcolor{red!15}0.029 & \cellcolor{red!15}0.024 \\
\hline
90ep & \cellcolor{red!45}4.0e-4 & \cellcolor{red!30}1.2e-3 & \cellcolor{red!15}0.022 & 0.16 & 0.29 & 0.93 \\
\hline
150ep & \cellcolor{red!45}6.6e-5 & \cellcolor{red!15}0.034 & \cellcolor{red!15}0.021 & 0.25 & 0.91 & 0.68 \\
\hline
300ep & \cellcolor{red!30}2.5e-3 & 0.86 & 0.84 & \cellcolor{red!15}0.013 & 0.36 & 0.14 \\
\hline
\end{tabular}
\caption{T-test p-values of the differences in top-1 val. accuracies (original) between the TensorFlow and PyTorch Inception crop, two-sided, equal variance, color-coded (\colorbox{red!15}{p \textless 0.05}, \colorbox{red!30}{p \textless 0.01}, \colorbox{red!45}{p \textless 0.001}).}
\label{table:t_test_tf_vs_torch_original}
\end{table}

\begin{table}[hbt!]
\small
\centering
\begin{tabular}{|c||c|c|c|c|c|c|}
\hline
$a_{min}$ & 0.025 & 0.05 & 0.1 & 0.15 & 0.2 & 0.25 \\
\hline
30ep & \cellcolor{red!45}2.3e-4 & \cellcolor{red!45}8.8e-4 & \cellcolor{red!15}0.01 & \cellcolor{red!15}0.024 & \cellcolor{red!30}1.6e-3 & \cellcolor{red!15}0.011 \\
\hline
60ep & \cellcolor{red!30}1.0e-3 & \cellcolor{red!45}5.7e-5 & \cellcolor{red!30}7.6e-3 & \cellcolor{red!15}0.027 & \cellcolor{red!15}0.034 & \cellcolor{red!30}2.4e-3 \\
\hline
90ep & \cellcolor{red!30}1.1e-3 & \cellcolor{red!30}7.2e-3 & \cellcolor{red!15}0.022 & 0.88 & 0.47 & 0.63 \\
\hline
150ep & \cellcolor{red!15}0.015 & 0.31 & \cellcolor{red!15}0.04 & 0.59 & 0.18 & 0.33 \\
\hline
300ep & 0.79 & 0.091 & \cellcolor{red!15}0.036 & \cellcolor{red!15}0.035 & 0.1 & 0.95 \\
\hline
\end{tabular}
\caption{T-test p-values of the differences in top-1 val. accuracies (ReaL) between the TensorFlow and PyTorch Inception crop, two-sided, equal variance, color-coded (\colorbox{red!15}{p \textless 0.05}, \colorbox{red!30}{p \textless 0.01}, \colorbox{red!45}{p \textless 0.001}).}
\label{table:t_test_tf_vs_torch_real}
\end{table}
 
\begin{table}[hbt!]
\small
\centering
\begin{tabular}{|c||c|c|c|c|c|c|}
\hline
$a_{min}$ & 0.025 & 0.05 & 0.1 & 0.15 & 0.2 & 0.25 \\
\hline
30ep & \cellcolor{red!45}1.8e-4 & \cellcolor{gray!50}1 & \cellcolor{red!30}2.1e-3 & \cellcolor{red!45}2.6e-5 & \cellcolor{red!45}3.e-7 & \cellcolor{red!45}2.5e-7 \\
\hline
60ep & \cellcolor{red!30}9.8e-3 & \cellcolor{gray!50}1 & \cellcolor{red!30}1.7e-3 & \cellcolor{red!45}7.1e-5 & \cellcolor{red!45}6.2e-5 & \cellcolor{red!45}4.4e-5 \\
\hline
90ep & \cellcolor{red!45}1.9e-4 & \cellcolor{gray!50}1 & \cellcolor{red!45}9.4e-4 & \cellcolor{red!45}1.5e-4 & \cellcolor{red!45}2.7e-4 & \cellcolor{red!45}7.0e-4 \\
\hline
150ep & \cellcolor{red!30}1.8e-3 & \cellcolor{gray!50}1 & \cellcolor{red!30}2.9e-3 & \cellcolor{red!30}9.9e-3 & \cellcolor{red!30}9.8e-3 & 0.31 \\
\hline
300ep & \cellcolor{red!30}5.e-3 & \cellcolor{gray!50}1 & 0.3 & \cellcolor{red!15}0.037 & \cellcolor{red!30}1.3e-3 & \cellcolor{red!30}1.2e-3 \\
\hline
\end{tabular}
\caption{T-test p-values of the top-1 val. accuracies (original) between the default $a_{min}=0.05$ and other values with various training budgets, TensorFlow Inception crop, two-sided, equal variance, color-coded (\colorbox{red!15}{p \textless 0.05}, \colorbox{red!30}{p \textless 0.01}, \colorbox{red!45}{p \textless 0.001}).}
\label{table:t_test_tf_trend_original}
\end{table}

\begin{table}[hbt!]
\small
\centering
\begin{tabular}{|c||c|c|c|c|c|c|}
\hline
$a_{min}$ & 0.025 & 0.05 & 0.1 & 0.15 & 0.2 & 0.25 \\
\hline
30ep & \cellcolor{red!30}1.5e-3 & \cellcolor{gray!50}1 & \cellcolor{red!30}3.4e-3 & \cellcolor{red!45}2.1e-4 & \cellcolor{red!45}7.4e-6 & \cellcolor{red!45}8.5e-6 \\
\hline
60ep & \cellcolor{red!15}0.013 & \cellcolor{gray!50}1 & \cellcolor{red!30}3.7e-3 & \cellcolor{red!45}3.4e-5 & \cellcolor{red!45}2.3e-5 & \cellcolor{red!45}5.2e-7 \\
\hline
90ep & \cellcolor{red!45}1.8e-4 & \cellcolor{gray!50}1 & \cellcolor{red!15}0.016 & \cellcolor{red!45}2.8e-4 & \cellcolor{red!45}3.1e-4 & \cellcolor{red!30}8.5e-3 \\
\hline
150ep & \cellcolor{red!15}0.013 & \cellcolor{gray!50}1 & 0.071 & 0.96 & 0.26 & 0.13 \\
\hline
300ep & 0.27 & \cellcolor{gray!50}1 & 0.16 & \cellcolor{red!30}4.6e-3 & \cellcolor{red!45}7.3e-4 & \cellcolor{red!30}1.4e-3 \\
\hline
\end{tabular}
\caption{T-test p-values of the top-1 val. accuracies (ReaL) between the default $a_{min}=0.05$ and other values with various training budget, TensorFlow Inception crop, two-sided, equal variance, color-coded (\colorbox{red!15}{p \textless 0.05}, \colorbox{red!30}{p \textless 0.01}, \colorbox{red!45}{p \textless 0.001}).}
\label{table:t_test_tf_trend_real}
\end{table}

\begin{table}[hbt!]
\small
\centering
\begin{tabular}{|c||c|c|c|c|c|c|}
\hline
$a_{min}$ & 0.025 & 0.05 & 0.1 & 0.15 & 0.2 & 0.25 \\
\hline
30ep & 0.43 & \cellcolor{gray!50}1 & \cellcolor{red!15}0.016 & \cellcolor{red!30}3.4e-3 & \cellcolor{red!45}7.3e-4 & \cellcolor{red!45}4.3e-4 \\
\hline
60ep & \cellcolor{red!30}2.7e-3 & \cellcolor{gray!50}1 & 0.38 & 0.081 & \cellcolor{red!30}2.1e-3 & \cellcolor{red!45}5.3e-4 \\
\hline
90ep & 0.39 & \cellcolor{gray!50}1 & 0.1 & 0.083 & \cellcolor{red!15}0.019 & 0.066 \\
\hline
150ep & 0.62 & \cellcolor{gray!50}1 & 0.34 & 0.6 & 0.42 & 0.17 \\
\hline
300ep & 0.45 & \cellcolor{gray!50}1 & 0.21 & \cellcolor{red!30}1.5e-3 & \cellcolor{red!30}1.2e-3 & \cellcolor{red!45}2.2e-4 \\
\hline
\end{tabular}
\caption{T-test p-values of the top-1 val. accuracies (original) between $a_{min}=0.05$ and other values with various training budgets, PyTorch Inception crop, two-sided, equal variance, color-coded (\colorbox{red!15}{p \textless 0.05}, \colorbox{red!30}{p \textless 0.01}, \colorbox{red!45}{p \textless 0.001}).}
\label{table:t_test_torch_trend_original}
\end{table}

\begin{table}[hbt!]
\small
\centering
\begin{tabular}{|c||c|c|c|c|c|c|}
\hline
$a_{min}$ & 0.025 & 0.05 & 0.1 & 0.15 & 0.2 & 0.25 \\
\hline
30ep & 0.27 & \cellcolor{gray!50}1 & \cellcolor{red!15}0.019 & \cellcolor{red!30}3.1e-3 & \cellcolor{red!45}7.4e-4 & \cellcolor{red!45}3.8e-4 \\
\hline
60ep & \cellcolor{red!15}0.023 & \cellcolor{gray!50}1 & 0.18 & 0.055 & \cellcolor{red!30}6.9e-3 & \cellcolor{red!45}4.3e-4 \\
\hline
90ep & 0.98 & \cellcolor{gray!50}1 & 0.057 & 0.17 & \cellcolor{red!15}0.042 & 0.19 \\
\hline
150ep & 0.6 & \cellcolor{gray!50}1 & 0.47 & 0.18 & \cellcolor{red!15}0.039 & \cellcolor{red!15}0.027 \\
\hline
300ep & 0.56 & \cellcolor{gray!50}1 & \cellcolor{red!45}2.6e-4 & \cellcolor{red!30}1.1e-3 & \cellcolor{red!45}4.9e-5 & \cellcolor{red!45}8.6e-5 \\
\hline
\end{tabular}
\caption{T-test p-values of the top-1 val. accuracies (ReaL) between $a_{min}=0.05$ and other values with various training budget, PyTorch Inception crop, two-sided, equal variance, color-coded (\colorbox{red!15}{p \textless 0.05}, \colorbox{red!30}{p \textless 0.01}, \colorbox{red!45}{p \textless 0.001}).}
\label{table:t_test_torch_trend_real}
\end{table}

\subsection{Comparison at equal augmentation strength}
\label{sec:inception_equal_strength_comparison}
\begin{figure*}
    \begin{minipage}{0.5\textwidth}
    \centering
    \includegraphics[width=0.8\textwidth]{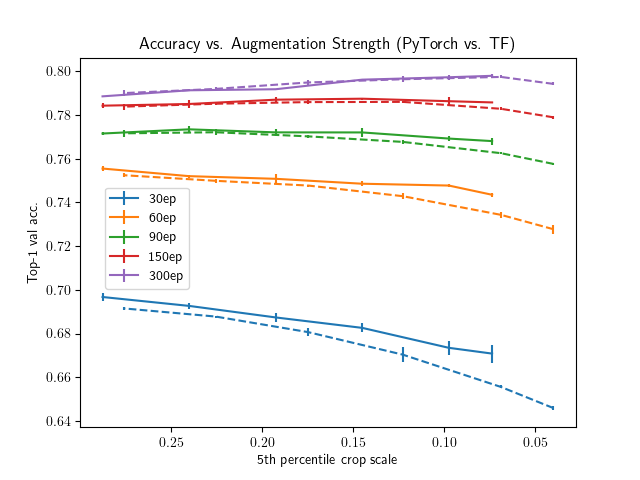}
    \end{minipage}%
    \begin{minipage}{0.5\textwidth}
    \centering
    \includegraphics[width=0.8\textwidth]{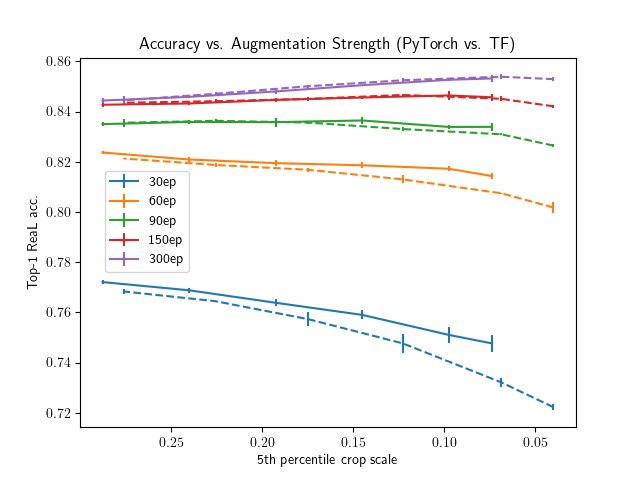}
    \end{minipage}%
    \caption{Top-1 val. accuracy of PyTorch Inception crop (solid line) vs. TensorFlow Inception crop (dashed line), original label (left) and ReaL (right), plotted with the 5th percentile crop scale as the x-axis. TensorFlow Inception crop generally underperforms except 300ep where it matches PyTorch Inception crop even with such unified augmentation strength measure.}
  \label{fig:torch_vs_tf_percentile}
\end{figure*}

Even though both are usually considered ``Inception crop'' and we compare them at the same crop scale lower bound $a_\mathrm{min}$, the difference in crop scale distribution between the TensorFlow and PyTorch Inception crop leaves the question open whether the discrepancy in model performance is merely due to different augmentation strength. \Cref{fig:torch_vs_tf_percentile} shows that TensorFlow Inception crop generally underperforms even when we use the 5th percentile crop scale as the unified measure of the augmentation strength, with reasonable interpolation.

\section{Detailed Beta crop results}
\label{sec:beta_full}
\subsection{Positive / Negative skew results with ReaL}
\begin{figure}
\centering
\begin{minipage}{0.5\textwidth}
    \raggedleft
    \small
    \resizebox{\linewidth}{!}{%
    \begin{tabular}{|c||c||c|c|c|}
    \hline
     & $a_{min}$ & 0.25 & 0.33 & 0.5 \\
    \hline
    \multirow{5}{*}{$\beta = 3$} & 30ep & 76.04$\pm 0.03$ & 76.84$\pm 0.17$ & \textbf{77.93}$\pm 0.15$ \\
    \cline{2-5}
     & 60ep & 81.75$\pm 0.02$ & 81.94$\pm 0.14$ & \textbf{82.14}$\pm 0.13$ \\
    \cline{2-5}
     & 90ep & 83.10$\pm 0.06$ & \textbf{83.17}$\pm 0.18$ & 82.99$\pm 0.16$ \\
    \cline{2-5}
     & 150ep & 83.98$\pm 0.21$ & \textbf{83.99}$\pm 0.07$ & 83.41$\pm 0.13$ \\
    \cline{2-5}
     & 300ep & \textbf{84.37}$\pm 0.04$ & 84.04$\pm 0.07$ & 83.41$\pm 0.11$ \\
    \hline
    \multirow{5}{*}{$\beta = 5$} & 30ep & 75.03$\pm 0.18$ & 76.46$\pm 0.21$ & \textbf{77.72}$\pm 0.23$ \\
    \cline{2-5}
     & 60ep & 81.05$\pm 0.04$ & 81.66$\pm 0.20$ & \textbf{82.20}$\pm 0.13$ \\
    \cline{2-5}
     & 90ep & 82.78$\pm 0.16$ & 82.99$\pm 0.12$ & \textbf{83.06}$\pm 0.14$ \\
    \cline{2-5}
     & 150ep & \textbf{83.92}$\pm 0.13$ & 83.79$\pm 0.04$ & 83.40$\pm 0.13$ \\
    \cline{2-5}
     & 300ep & \textbf{84.13}$\pm 0.14$ & 84.02$\pm 0.08$ & 83.35$\pm 0.15$ \\
    \hline
    \end{tabular}}
\end{minipage}
\hfill
\begin{minipage}{0.4\textwidth}
    \raggedright
    \includegraphics[width=\textwidth]{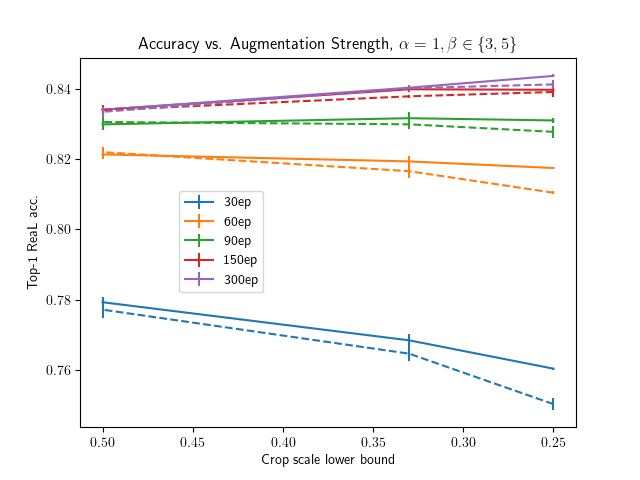}
\end{minipage}
\caption{Top-1 val. accuracy (ReaL), Beta crop with $\alpha=1$. Solid line: $\beta=3$, Dashed line: $\beta=5$}
\label{fig:beta_real_full}
\end{figure}

\begin{figure}
\centering
\begin{minipage}{0.5\textwidth}
    \raggedleft
    \small
    \begin{tabular}{|c||c||c|c|}
    \hline
     & $a_{min}$ & 0.05 & 0.1 \\
    \hline
    \multirow{5}{*}{$\alpha = 2$} & 30ep & 77.28$\pm 0.11$ & \textbf{77.67}$\pm 0.18$ \\
    \cline{2-4}
     & 60ep & 82.28$\pm 0.16$ & \textbf{82.34}$\pm 0.10$ \\
    \cline{2-4}
     & 90ep & 83.60$\pm 0.27$ & \textbf{83.61}$\pm 0.01$ \\
    \cline{2-4}
     & 150ep & \textbf{84.43}$\pm 0.08$ & 84.20$\pm 0.15$ \\
    \cline{2-4}
     & 300ep & \textbf{84.67}$\pm 0.21$ & 84.41$\pm 0.04$ \\
    \hline
    \multirow{5}{*}{$\alpha = 3$} & 30ep & 77.79$\pm 0.17$ & \textbf{77.86}$\pm 0.18$ \\
    \cline{2-4}
     & 60ep & \textbf{82.35}$\pm 0.18$ & 82.22$\pm 0.14$ \\
    \cline{2-4}
     & 90ep & 83.37$\pm 0.02$ & \textbf{83.44}$\pm 0.09$ \\
    \cline{2-4}
     & 150ep & \textbf{84.04}$\pm 0.14$ & 83.91$\pm 0.15$ \\
    \cline{2-4}
     & 300ep & \textbf{84.12}$\pm 0.15$ & 83.88$\pm 0.13$ \\
    \hline
    \multirow{5}{*}{$\alpha = 5$} & 30ep & 78.00$\pm 0.16$ & \textbf{78.17}$\pm 0.21$ \\
    \cline{2-4}
     & 60ep & 82.01$\pm 0.20$ & \textbf{82.09}$\pm 0.01$ \\
    \cline{2-4}
     & 90ep & \textbf{83.04}$\pm 0.09$ & 82.85$\pm 0.11$ \\
    \cline{2-4}
     & 150ep & \textbf{83.48}$\pm 0.16$ & 83.32$\pm 0.12$ \\
    \cline{2-4}
     & 300ep & \textbf{83.30}$\pm 0.24$ & 83.19$\pm 0.13$ \\
    \hline
    \end{tabular}
\end{minipage}
\hfill
\begin{minipage}{0.4\textwidth}
    \raggedright
    \includegraphics[width=\linewidth]{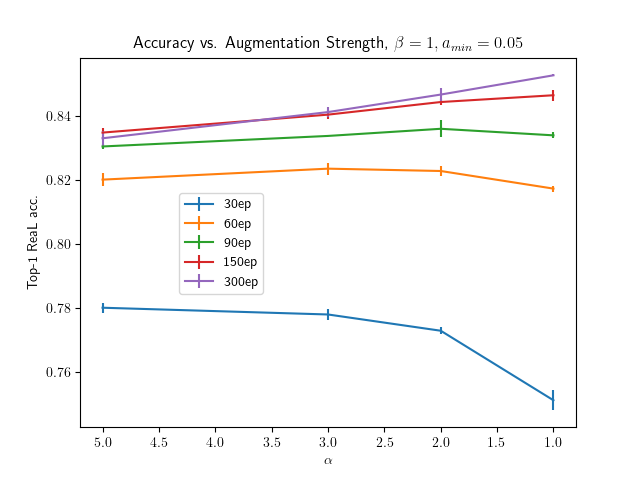}
\end{minipage}
\caption{Top-1 val. accuracy (ReaL), Beta crop with $\beta=1$. Top-1 val. accuracy vs. $\alpha$, $a_{min} = 0.05$ is plotted on the right. For $\alpha=\beta=1$, we reuse the data points of the PyTorch Inception crop experiments.}
\label{fig:alpha_val_full_real}
\end{figure}

\Cref{fig:beta_real_full,fig:alpha_val_full_real} show the ReaL evaluation results of the ViT-S/16 models trained on ImageNet-1k with Beta crop, positive / negative skew. We train $N=3$ for each setting and values are given as (mean) $\pm$ (sample standard deviation), both rounded to 2 decimal places.

\subsection{Matching strength, full tables with error bars}
\Cref{table:matching_strength_original_full,table:matching_strength_real_full} show the full evaluation results of the ViT-S/16 models trained on ImageNet-1k with Beta crop, matching strength, original label (\cref{table:matching_strength_original_full}) and ReaL (\cref{table:matching_strength_real_full}). We train $N=3$ for each setting and values are given as (mean) $\pm$ (sample standard deviation), both rounded to 2 decimal places.

\begin{table*}[hbt!]
\small
\centering
\begin{tabular}{|c||c|c|c|c|c|c|c|c|c|c|c|c|}
\hline
$\alpha$ & 1.149 & 1.287 & 1.551 & 1.818 & 2.099 & 2.403 \\
\hline
30ep & 67.47$\pm 0.18$ & 67.92$\pm 0.20$ & 68.67$\pm 0.15$ & 69.22$\pm 0.25$ & 69.61$\pm 0.19$ & \textbf{69.87}$\pm 0.04$ \\
\hline
60ep & 74.59$\pm 0.13$ & 74.79$\pm 0.14$ & 75.24$\pm 0.17$ & 75.32$\pm 0.12$ & 75.48$\pm 0.15$ & \textbf{75.51}$\pm 0.04$ \\
\hline
90ep & 76.89$\pm 0.04$ & 76.99$\pm 0.23$ & \textbf{77.39}$\pm 0.11$ & 77.25$\pm 0.15$ & 77.15$\pm 0.14$ & 77.29$\pm 0.21$ \\
\hline
150ep & 78.63$\pm 0.21$ & 78.49$\pm 0.07$ & \textbf{78.73}$\pm 0.06$ & 78.59$\pm 0.24$ & 78.63$\pm 0.09$ & 78.59$\pm 0.12$ \\
\hline
300ep & 79.57$\pm 0.10$ & \textbf{79.60}$\pm 0.11$ & 79.42$\pm 0.10$ & 79.27$\pm 0.15$ & 78.95$\pm 0.23$ & 79.04$\pm 0.12$ \\
\hline
\end{tabular}
\caption{Top-1 val. accuracy of Beta crop, matching strength (original label). As the training budget increases from 30 epochs to 300 epochs, the optimal $\alpha$ decreases (stronger augmentation). $\alpha = 1.149$ may be an exception possibly due to too many crops near zero area. In future iterations we may want small but nonzero $a_{\min}$ (say $0 < a_{\min} \le 0.025$) instead while keeping the same 5th percentile crop scale.}
\label{table:matching_strength_original_full}
\end{table*}

\begin{table*}[hbt!]
\small
\centering
\begin{tabular}{|c||c|c|c|c|c|c|c|c|c|c|c|c|}
\hline
$\alpha$ & 1.149 & 1.287 & 1.551 & 1.818 & 2.099 & 2.403 \\
\hline
30ep & 75.16$\pm 0.15$ & 75.58$\pm 0.17$ & 76.44$\pm 0.12$ & 76.97$\pm 0.23$ & 77.30$\pm 0.08$ & \textbf{77.58}$\pm 0.10$ \\
\hline
60ep & 81.72$\pm 0.05$ & 81.85$\pm 0.06$ & 82.20$\pm 0.10$ & 82.17$\pm 0.16$ & \textbf{82.35}$\pm 0.15$ & 82.35$\pm 0.05$ \\
\hline
90ep & 83.48$\pm 0.04$ & 83.56$\pm 0.11$ & \textbf{83.72}$\pm 0.09$ & 83.59$\pm 0.06$ & 83.54$\pm 0.17$ & 83.72$\pm 0.15$ \\
\hline
150ep & \textbf{84.72}$\pm 0.16$ & 84.48$\pm 0.08$ & 84.63$\pm 0.03$ & 84.48$\pm 0.10$ & 84.41$\pm 0.08$ & 84.42$\pm 0.06$ \\
\hline
300ep & \textbf{85.15}$\pm 0.16$ & 85.14$\pm 0.03$ & 84.90$\pm 0.10$ & 84.77$\pm 0.16$ & 84.58$\pm 0.11$ & 84.55$\pm 0.14$ \\
\hline
\end{tabular}
\caption{Rop-1 val. accuracy of Beta crop, matching strength (ReaL). As the training budget increases from 30 epochs to 300 epochs, the optimal $\alpha$ decreases (stronger augmentation).}
\label{table:matching_strength_real_full}
\end{table*}

\section{More plots for gradient, Grad-CAM, and relevance entropy}
\subsection{Gradient entropy}
\label{sec:grad_entropy_plots}
\begin{figure*}
    \centering
    \includegraphics[width=\textwidth]{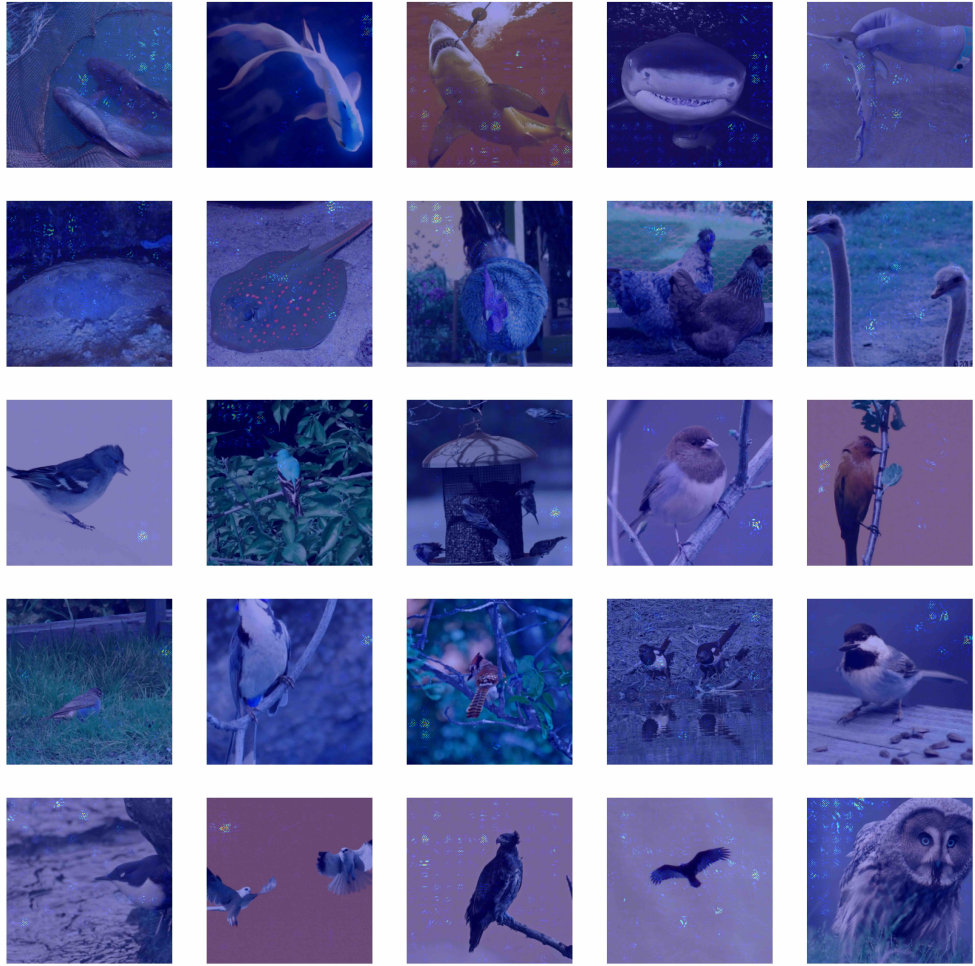}
    \caption{Sample saliency maps given by the rectified gradient for the 1st images of the first 25 classes $c \in [0\ldotp \ldotp 24]$ in \texttt{torchvision.datasets.ImageNet} validation split. We use a model trained with 300ep training budget and the TensorFlow Inception crop with $a_\mathrm{min}=0.05$. Though conceptually the simplest, the saliency maps given by the rectified gradient are the least interpretable of the 3.}
    \label{fig:grad_first_25_classes}
\end{figure*}

\begin{figure}[H]
  \centering
  \includegraphics[width=0.4\textwidth]{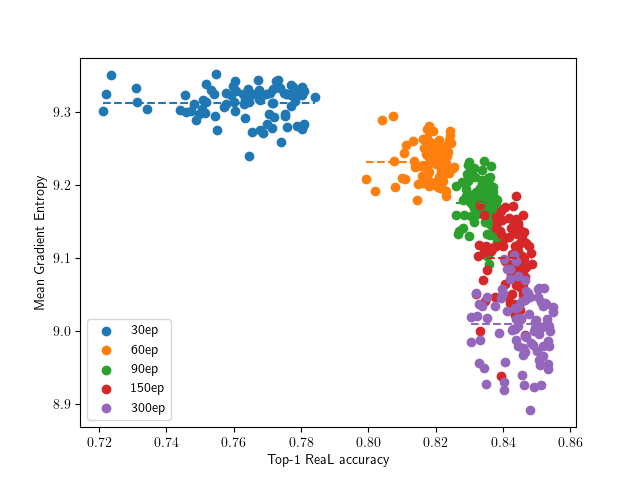}
  \caption{Mean gradient entropy $\overline{S_\mathrm{grad}}$ defined in \cref{eq:grad_entropy} vs. top-1 val. accuracy (ReaL) for all 450 Inception crop and Beta crop experiments with AdamW optimizer, color-coded by $a_\mathrm{min}$.}
  \label{fig:grad_entropy_amin_real}
\end{figure}

\begin{figure}[H]
    \centering
    \begin{minipage}{0.4\textwidth}
    \centering
    \includegraphics[width=\textwidth]{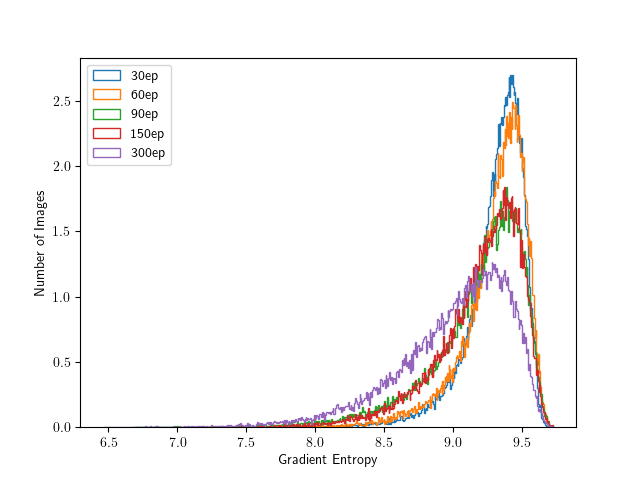}
    \end{minipage}
    \begin{minipage}{0.4\textwidth}
    \centering
    \includegraphics[width=\textwidth]{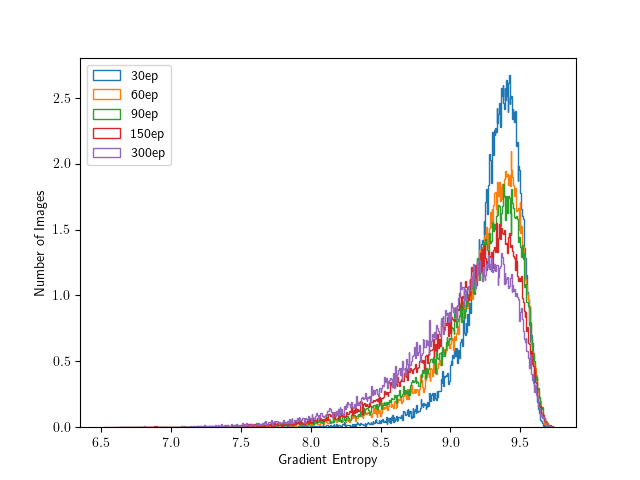}
    \end{minipage}%
    \caption{Distribution of the gradient entropy $S_\mathrm{grad}$ defined in \cref{eq:grad_entropy} for the 50000 validation set images given by models trained with the TensorFlow Inception crop (left) and PyTorch Inception crop (right) with $a_\mathrm{min} = 0.05$, color-coded by the training budget.}
    \label{fig:grad_entropy_distribution}
\end{figure}

\Cref{fig:grad_first_25_classes} shows some sample saliency maps given by the rectified gradient. Plotting the mean gradient entropy against the top-1 validation accuracy using the multi-label ReaL does not change the whole picture: The mean gradient entropy is still negatively correlated with the training budget (\cref{fig:grad_entropy_amin_real}) as the distribution of $S_\mathrm{grad}$ for the 50000 validation set images shifts to the left (\cref{fig:grad_entropy_distribution}).

\subsection{Grad-CAM entropy}
\label{sec:grad_cam_entropy_plots}
\begin{figure*}
    \centering
    \includegraphics[width=\textwidth]{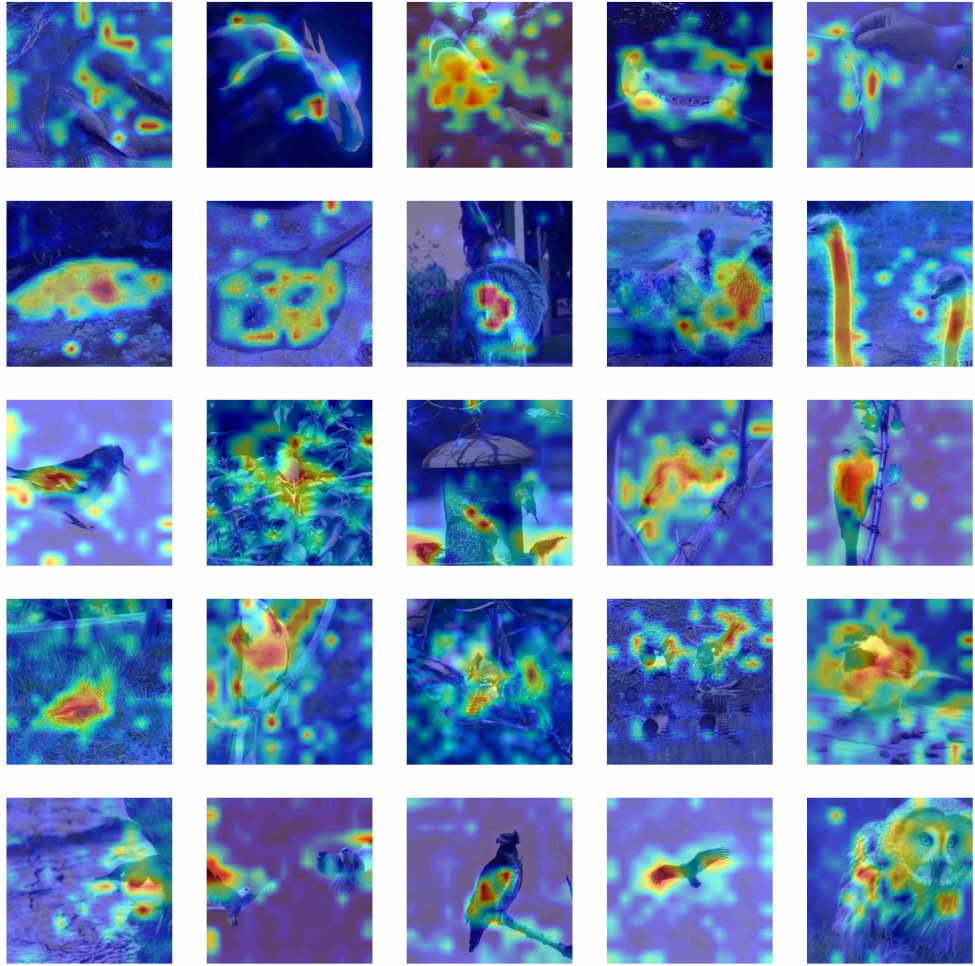}
    \caption{Sample saliency maps given by Grad-CAM at the last layer for the 1st images of the first 25 classes $c \in [0\ldotp \ldotp 24]$ in \texttt{torchvision.datasets.ImageNet} validation split. We use a model trained with 300ep training budget and the TensorFlow Inception crop with $a_\mathrm{min}=0.05$.}
    \label{fig:grad_cam_last_layer_first_25_classes}
\end{figure*}

\begin{figure}
    \centering
    \begin{minipage}{0.4\textwidth}
    \centering
    \includegraphics[width=\textwidth]{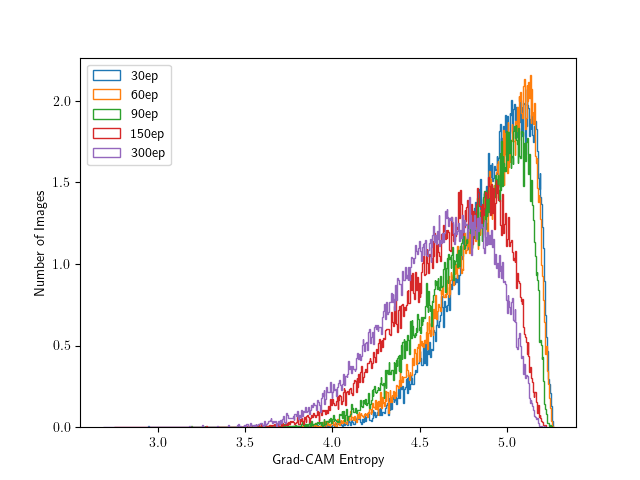}
    \end{minipage}
    \begin{minipage}{0.4\textwidth}
    \centering
    \includegraphics[width=\textwidth]{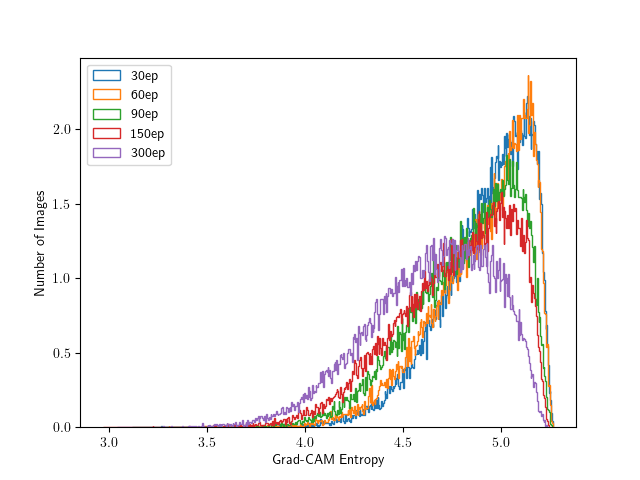}
    \end{minipage}%
    \caption{Distribution of the Grad-CAM entropy $S_\mathrm{grad-cam}$ for the 50000 validation set images given by models trained with the TensorFlow Inception crop (left) and PyTorch Inception crop (right) with $a_\mathrm{min} = 0.05$, color-coded by the training budget.}
    \label{fig:grad_cam_entropy_distribution}
\end{figure}

\Cref{fig:grad_cam_last_layer_first_25_classes} shows some sample saliency maps given by Grad-CAM of the last layer and \cref{fig:grad_cam_entropy_distribution} shows the distribution of the Grad-CAM entropy $S_\mathrm{grad-cam}$ for the 50000 validation set images with various training budgets while keeping $a_\mathrm{min} = 0.05$ fixed. 

\subsection{Relevance entropy}
\label{sec:rel_entropy_plots}
\begin{figure*}
    \centering
    \includegraphics[width=\textwidth]{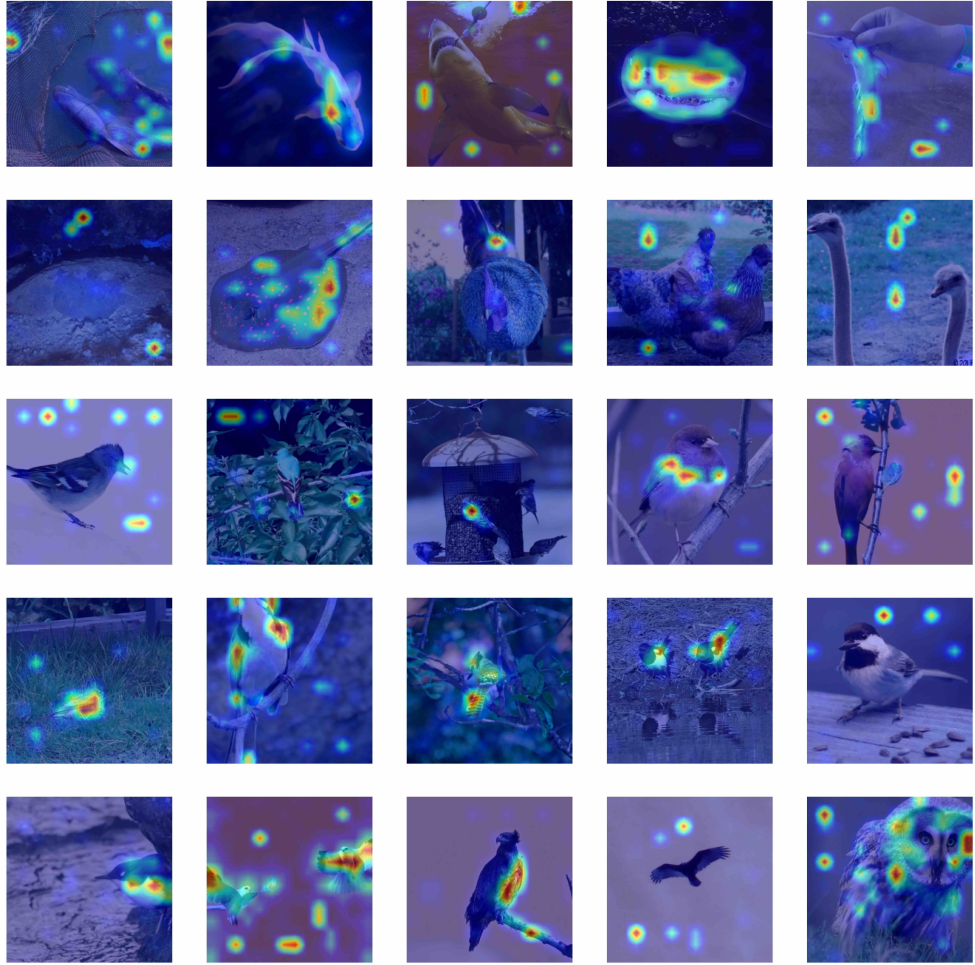}
    \caption{Sample saliency maps given by LRP at the last layer for the 1st images of the first 25 classes $c \in [0\ldotp \ldotp 24]$ in \texttt{torchvision.datasets.ImageNet} validation split. We use a model trained with 300ep training budget and the TensorFlow Inception crop with $a_\mathrm{min}=0.05$.}
    \label{fig:lrp_last_layer_first_25_classes}
\end{figure*}

\begin{figure}
    \centering
    \begin{minipage}{0.4\textwidth}
    \centering
    \includegraphics[width=\textwidth]{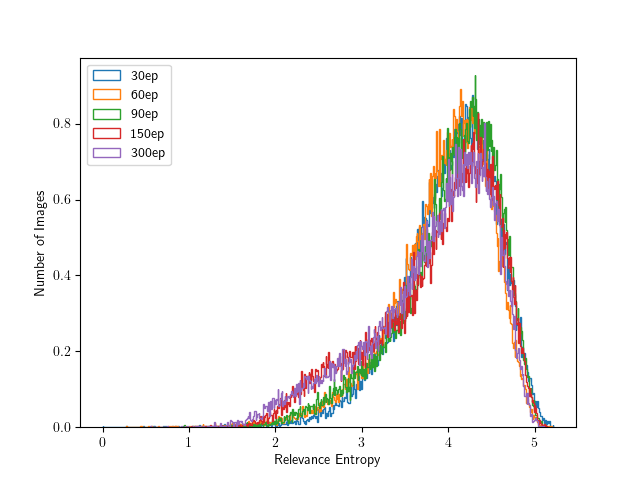}
    \end{minipage}
    \begin{minipage}{0.4\textwidth}
    \centering
    \includegraphics[width=\textwidth]{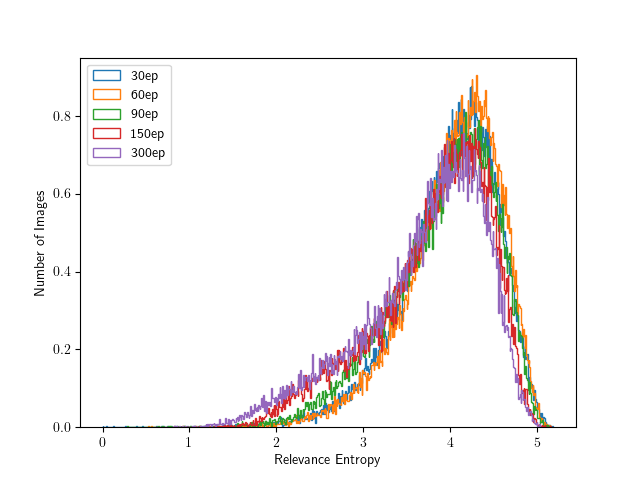}
    \end{minipage}%
    \caption{Distribution of the relevance entropy $S_\mathrm{rel}$ for the 50000 validation set images given by models trained with the TensorFlow Inception crop (left) and PyTorch Inception crop (right) with $a_\mathrm{min} = 0.05$, color-coded by the training budget.}
    \label{fig:rel_entropy_distribution}
\end{figure}

\Cref{fig:lrp_last_layer_first_25_classes} shows some sample saliency maps given by LRP of the last layer and \cref{fig:rel_entropy_distribution} shows the distribution of the relevance entropy $S_\mathrm{rel}$ for the 50000 validation set images with various training budgets while keeping $a_\mathrm{min} = 0.05$ fixed. Compared to the gradient entropy counterpart (\cref{fig:grad_entropy_distribution}) the sparsification at the last layer is more modest, but the trend remains: The higher the training budget, the sparser the attention becomes.

\section{Scion experiment setup and WD sweep}
\label{sec:scion_setup}

The constrained variant of Scion \citep{pethick2025training} can be considered a collection of optimizers with the following unified update rules. Given loss function $f_t$ at time $t$, layer $l$ and layer weight $\bm{\theta}_{t,l}$ at time $t-1$, the choice of linear minimization oracle $\lmo_l$, momentum $\alpha$, learning rate $\gamma$, and radius $\rho_l$:
\begin{align*}
\vg_{t,l} &\leftarrow \nabla_{\theta_l} f_t (\bm{\theta}_{t-1,l}, \zeta_t) \\
\vm_{t,l} &\leftarrow (1 - \alpha) \vm_{t-1,l} + \alpha \vg_{t,l} \\
\bm{\theta}_{t,l} &\leftarrow (1 - \gamma)\bm{\theta}_{t-1, l} + \gamma \rho_l \lmo_l(\vm_{t,l})
\end{align*}
\Cref{tbl:norms_lmos} lists the $\lmo$s and the norms from which they are derived that we use in our experiments. Conceptually, we choose the norms of the layers based on the shape of the weight and their functions in the model, and $\lmo$s are the updates with unit norms in the direction of the steepest descent.

Although equivalent up to reparameterization, it is difficult to perform a WD sweep or the equivalent in the original formulation of Scion. We therefore reformulate constrained Scion in terms of independent weight decay coefficient $\eta = \gamma$, layer-wise learning rate $\gamma_l = \gamma \rho_l$, and layer-wise weight decay coefficient $\lambda_l = \frac{1}{\rho_l}$. The update rules then become
\begin{align*}
\vg_{t,l} &\leftarrow \nabla_{\theta_l} f_t (\bm{\theta}_{t-1,l}, \zeta_{t}) \\
\vm_{t,l} &\leftarrow (1 - \alpha) \vm_{t-1,l} + \alpha \vg_{t,l} \\
\bm{\theta}_{t,l} &\leftarrow (1 - \textcolor{red}{\eta})\bm{\theta}_{t-1, l} + \textcolor{red}{\gamma_l} \lmo_l(\vm_{t,l}) \\
&= \bm{\theta}_{t-1,l} + \textcolor{red}{\gamma_l} \left(\textcolor{red}{-\lambda_l} \bm{\theta}_{t-1,l} + \lmo_l(\vm_{t,l}) \right)
\end{align*}

\begin{table*}
\centering
\caption{Norms and the associated $\lmo$s as normalized in our experiments. Sign and Spectral assume matrix weight $\bm{\theta}_{l} = \mA \in \R^{d_\mathrm{out} \times d_\mathrm{in}}$ while Bias assumes vector weight $\bm{\theta}_{l} = \vb_\ell \in \R^{d_\mathrm{out}}$. $\mU \mV^\top$ refers to the reduced SVD of the input matrix with unitary matrices $\mU$ and $\mV^\top$ from the full SVD $\mA=\mU\text{diag}(\bm{\sigma}) \mV^\top$ while $\|\mA\|_{\mathcal{S}_{\infty}} = \max(\bm{\sigma})$ is the spectral norm of the matrix.}
\label{tbl:norms_lmos}
\bgroup
\def\arraystretch{1.2}
\begin{tabular}{|c|c|c|c|}
\hline
& Sign & Spectral & Bias \\
\hline\hline
\textbf{Norm} & $d_\mathrm{in} \max_{i,j} |A_{i,j}|$ & $\sqrt{\frac{d_\mathrm{in}}{d_\mathrm{out}}}\|\mA\|_{\mathcal{S}_{\infty}}$ & $\RMS$ \\
\hline
\textbf{LMO} & $\mA\mapsto -\frac{\sign(\mA)}{d_\mathrm{in}}$ & $\mA\mapsto-\sqrt{\frac{d_\mathrm{out}}{d_\mathrm{in}}}\mU\mV^\top$ & $\vb_\ell \mapsto -\tfrac{\vb_\ell}{\|\vb_\ell\|_\RMS}$ \\
\hline
\end{tabular}
\egroup
\end{table*}

\begin{figure}
    \centering
    \begin{minipage}{0.4\textwidth}
    \centering
    \includegraphics[width=\textwidth]{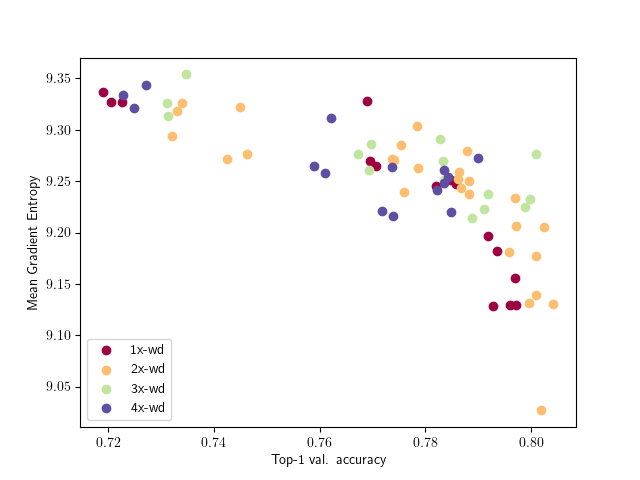}
    \end{minipage}
    \begin{minipage}{0.4\textwidth}
    \centering
    \includegraphics[width=\textwidth]{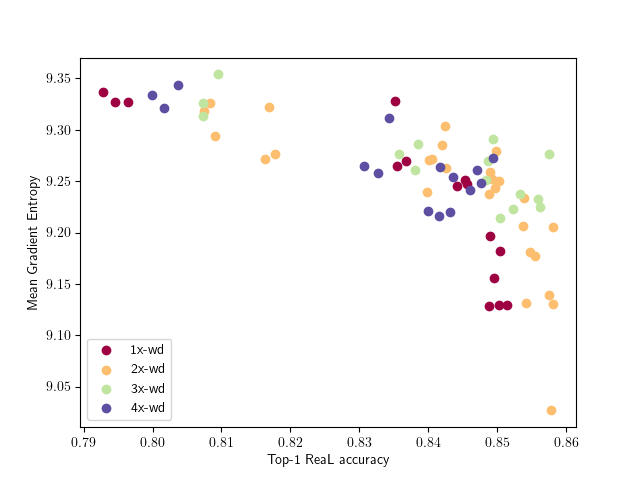}
    \end{minipage}%
    \caption{Mean gradient entropy $\overline{S_\mathrm{grad}}$ vs. top-1 val. accuracy, original label (left) and ReaL (right), for all 72 experiments with Scion optimizer, color-coded by $\lambda_l$.}
    \label{fig:scion_grad_entropy_wd}
\end{figure}

We then adopt the following modifications from \citet{pethick2025training} for our Simple ViT-S/16:
\begin{enumerate}
  \item Scale the GELU activation function as $\sqrt{2}$GELU to preserve variance.
  \item Replace LayerNorm with RMSNorm.
\end{enumerate}
We also keep its batch size 4096, cosine learning rate decay with no warm-up, choices of $\lmo$s (Spectral for the input patchifier and hidden layers, Sign for the output layer, and Bias for the biases) and maximum learning rates ($\gamma_L = \gamma \rho_L = 0.0004 \times 500 = 0.2$ for the output Sign layer and $\gamma_l = \gamma \rho_l = 0.0004 \times 25 = 0.01$ for the rest). The WD coefficient (converted to our formulation) from \citet{pethick2025training} turns out to be suboptimal so we end up doing our own sweep with $\lambda_l \in \{0.04, \mathbf{0.08}, 0.12, 0.16\}$ and $\lambda_L = 0.05\lambda_l$ for the output Sign layer. The best value $\lambda_l = 0.08$ (\cref{table:scion_wd}) is used for the comparison in \cref{sec:scion_rep}. Including the optimized $a_{\min}$ experiments (\cref{table:scion_wd}), we have a total of 72 models trained with Scion optimizer. Plotting their mean gradient entropy $\overline{S_\mathrm{grad}}$ defined in \cref{eq:grad_entropy} vs. top-1 val. accuracy, original label (left) and ReaL (right), for all 72 experiments with Scion optimizer, color-coded by $\lambda_l$ shows no correlation between $\overline{S_\mathrm{grad}}$ and WD (\cref{fig:scion_grad_entropy_wd}).

\begin{table*}[hbt!]
\small
\centering
\begin{tabular}{|c||c|c|c|c|c|c|c|c|c|c|c|c|}
\hline
& \multicolumn{4}{c|}{Original label} & \multicolumn{4}{c|}{ReaL} \\
\hline \hline
$\lambda_l$ & 0.04 & 0.08 & 0.12 & 0.16 & 0.04 & 0.08 & 0.12 & 0.16 \\
\hline
30ep & 72.08$\pm 0.19$ & \textbf{73.31}$\pm 0.09$ & 73.24$\pm 0.21$ & 72.50$\pm 0.22$ & 79.46$\pm 0.18$ & \textbf{80.83}$\pm 0.07$ & 80.81$\pm 0.13$ & 80.18$\pm 0.20$ \\
\hline
60ep & 76.98$\pm 0.09$ & \textbf{77.44}$\pm 0.09$ & 76.88$\pm 0.14$ & 76.07$\pm 0.16$ & 83.58$\pm 0.08$ & \textbf{84.09}$\pm 0.10$ & 83.75$\pm 0.15$ & 83.26$\pm 0.18$ \\
\hline
90ep & 78.42$\pm 0.19$ & \textbf{78.68}$\pm 0.09$ & 78.32$\pm 0.04$ & 77.31$\pm 0.12$ & 84.51$\pm 0.08$ & \textbf{84.94}$\pm 0.05$ & 84.88$\pm 0.05$ & 84.11$\pm 0.10$ \\
\hline
150ep & 79.27$\pm 0.08$ & \textbf{79.65}$\pm 0.07$ & 79.05$\pm 0.15$ & 78.31$\pm 0.07$ & 84.94$\pm 0.09$ & \textbf{85.41}$\pm 0.05$ & 85.20$\pm 0.14$ & 84.69$\pm 0.08$ \\
\hline
300ep & 79.67$\pm 0.06$ & \textbf{80.10}$\pm 0.14$ & 79.98$\pm 0.10$ & 78.64$\pm 0.32$ & 85.04$\pm 0.10$ & \textbf{85.66}$\pm 0.21$ & 85.66$\pm 0.09$ & 84.54$\pm 0.35$ \\
\hline
\end{tabular}
\caption{Top-1 val. accuracy of the Scion optimizer WD sweep, original label (left) and ReaL (right).}
\label{table:scion_wd}
\end{table*}

\end{document}